\pdfoutput=1
\documentclass[11pt]{article}

\usepackage[final]{acl}
\usepackage[linesnumbered,ruled,vlined]{algorithm2e}
\usepackage{algpseudocode}
\usepackage{times}
\usepackage{latexsym}
\usepackage{wrapfig}

\usepackage[utf8]{inputenc}
\DeclareUnicodeCharacter{2009}{\,}
\DeclareUnicodeCharacter{2248}{\,}
\usepackage[T1]{fontenc}
\usepackage[utf8]{inputenc}

\usepackage{microtype}

\usepackage{inconsolata}

\usepackage{graphics}
\usepackage{amsmath}
\usepackage{booktabs}
\usepackage{multirow}
\usepackage{graphicx}
\usepackage{todonotes}

\title{FairCoT: Enhancing Fairness in Text-to-Image Generation via Chain of Thought Reasoning with Multimodal Large Language Models}
\author{
  Zahraa Al Sahili\textsuperscript{1}\thanks{\ \ Corresponding author.} \quad
  Ioannis Patras\textsuperscript{1} \quad
  Matthew Purver\textsuperscript{1,2} \\
  \textsuperscript{1} School of Electronic Engineering and Computer Science, \\ Queen Mary University of London, London, UK \\
  \textsuperscript{2} Department of Knowledge Technologies, \\ Jožef Stefan Institute, Ljubljana, Slovenia \\
  \texttt{\{z.alsahili, i.patras, m.purver\}@qmul.ac.uk}
}

\begin{document}
\maketitle
\begin{abstract}
    In the domain of text-to-image generative models, biases inherent in training datasets often propagate into generated content, posing significant ethical challenges, particularly in socially sensitive contexts. We introduce FairCoT, a novel framework that enhances fairness in text-to-image models through Chain-of-Thought (CoT) reasoning within multimodal generative large language models. FairCoT employs iterative CoT refinement to systematically mitigate biases, and dynamically adjusts textual prompts in real time, ensuring diverse and equitable representation in generated images. By integrating iterative reasoning processes, FairCoT addresses the limitations of zero-shot CoT in sensitive scenarios, balancing creativity with ethical responsibility. Experimental evaluations across popular text-to-image systems—including DALL-E and various Stable Diffusion variants—demonstrate that FairCoT significantly enhances fairness and diversity without sacrificing image quality or semantic fidelity. By combining robust reasoning, lightweight deployment, and extensibility to multiple models, FairCoT represents a promising step toward more socially responsible and transparent AI-driven content generation. Code, data, and prompt templates are available at: https://github.com/zahraaalsahili/FairCoT.

\end{abstract}

\section{Introduction}

Recent advances in text-to-image (T2I) generation have received significant attention in the Natural Language Processing and Computer Vision communities. By leveraging large text corpora and multimodal data, modern generative models can produce high-fidelity, contextually relevant images from textual descriptions. These breakthroughs have enabled new applications in creative content generation, design prototyping, and accessibility tools for language-driven image creation. However, the data-intensive nature of these models also risks propagating societal biases inherent in large-scale textual and visual datasets, potentially resulting in outputs that are biased or discriminatory. Addressing these biases is a critical challenge in text-to-image generation, and leveraging the reasoning capabilities of multimodal large language models offers promising avenues for mitigating potential discriminatory outputs.

\begin{figure}[ht]
\centering
\includegraphics[width=0.4\textwidth]{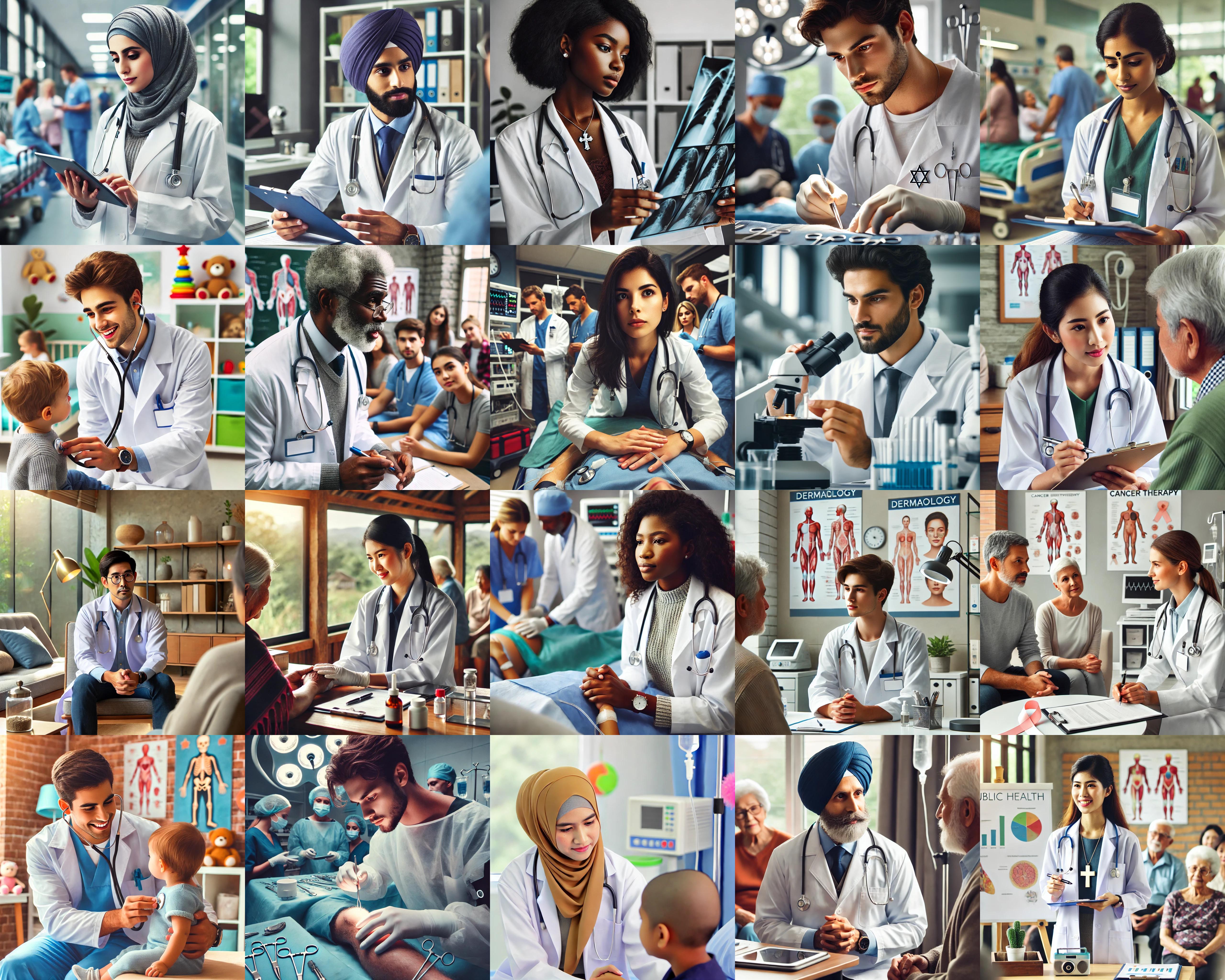} % Adjust width as necessary
\caption{Generated using DALL-E with FairCoT inference, this collection of images depicts doctors in diverse settings, showcasing a wide range of gender, race, age, and religious backgrounds. }
\label{fig:imgx}
\end{figure}

Efforts to mitigate bias in text-conditioned image generation typically focus on two broad categories: prompt engineering and model-based debiasing. On the one hand, prompt engineering strategies \citep{biasp1,biasp2} rely on human-authored templates, which are prone to subjectivity and require substantial manual labor \citep{sun2023enhancing}. On the other hand, fine-tuning or text-embedding modification approaches \citep{biasp3,finetuning,5} can be computationally expensive and often limited to specific open-source diffusion frameworks. Moreover, such model-centric solutions can inadvertently hamper alignment goals or overlook certain categories of bias if they were not explicitly addressed during the fine-tuning phase \citep{sun2023enhancing}. As the generative modeling ecosystem expands to include both open and closed-source platforms, there is a pressing need for lightweight and generalizable methods that can dynamically handle multiple forms of bias without costly retraining.

In this paper, we introduce \textbf{FairCoT}, a novel framework that incorporates \emph{Chain-of-Thought (CoT) reasoning} within \emph{multimodal large language models (MLLMs)} to refine and guide the generative process of T2I models toward more equitable outputs. Unlike zero-shot CoT approaches that may overlook complex social contexts, \textbf{FairCoT} adopts an \emph{iterative refinement} mechanism to produce \emph{fairness-aware} reasoning paths. It begins by assessing potential biases—such as those related to gender, race, age, or religion—and then adjusts textual prompts accordingly to ensure diverse and inclusive image generation. During iterative reasoning, the model updates its textual guidance based on intermediate feedback, thereby adapting to socially sensitive attributes in real time.

Additionally, we show how the method is flexible enough to incorporate multiple attributes simultaneously; here, proposing an \emph{attire-based} attribute detection strategy that leverages CLIP to expand the scope of bias identification to include religious attire and similar \emph{subtle} signals. This approach facilitates more accurate detection of sensitive attributes beyond commonly studied demographic categories. Lastly, we enable \emph{task-adaptive} CoT inference --- at inference time, the MLLM selects the most relevant chain of reasoning for generating fairness-conscious prompts, functioning as a lightweight plug-in that does not require retraining or parameter updates to underlying diffusion models, making it generalizable to biases beyond professions,including animal breeds, devices, and modes of transport (Figure~\ref{fig:imgx}).

Unlike traditional prompt engineering or fine-tuning methods, \textbf{FairCoT} operates at the \emph{reasoning level}, making it compatible with both open and closed-source T2I models. Our experiments—spanning popular systems like DALL-E and Stable Diffusion—demonstrate improved \emph{fairness and diversity} metrics without compromising \emph{image quality} or \emph{semantic relevance}. By offering a transparent and dynamic mechanism for controlling how T2I models respond to sensitive inputs, \textbf{FairCoT} represents a step toward more accountable multimodal systems, giving users on-demand options to address potential over- or underrepresentation of minority groups in specific contexts. Our contributions are summarized as follows:
\begin{enumerate}
    \item \textbf{FairCoT Framework:} We propose a Chain-of-Thought-based bias mitigation pipeline that seamlessly integrates with existing T2I models, avoiding parameter modifications and preserving alignment.
    \item \textbf{Multi-Bias Generalization:} \textbf{FairCoT} handles a range of biases—spanning sensitive attributes (e.g., race, gender, religion) and object-centric disparities—and also generalizes beyond professions (e.g., scenarios involving children, animals, or diverse objects) and  single-person settings, thus demonstrating broad applicability across domains.
    \item \textbf{Improved Attribute Prediction:} We develop an \emph{attire-focused} detection method leveraging CLIP to identify religiously contextual attributes—marking one of the first attempts to systematically address religious bias in T2I models.
    \item \textbf{Reasoning-Based Debiasing with MLLMs:} We utilize an \emph{iterative CoT} process that harnesses the reasoning capabilities of MLLMs to debias T2I models—particularly those with limited input token sizes and lacking inherent reasoning—to provide step-by-step fairness guidance for improved bias mitigation.
\end{enumerate}

The remainder of this paper is organized as follows: 
Section~\ref{sec:related_work} reviews related work on bias mitigation in generative modeling. 
Section~\ref{sec:methodology} details the \textbf{FairCoT} architecture, including its iterative CoT generation and inference components. 
Section~\ref{sec:experiments} presents our experimental setup and results, 
followed by a broader discussion of ethical implications and future work in Section~\ref{sec:discussion}.

\section{Related Work}
\label{sec:related_work}
This section discusses relevant work on bias in vision-language models, language-driven bias mitigation, advances in large language models, and attribute prediction.

\paragraph{Bias in Vision-Language Models}

Bias in vision-language models is a significant concern, as these models often inherit and amplify societal biases present in training data. \citet{6} investigated gender bias in zero-shot vision-language models, revealing performance disparities and calibration issues based on perceived gender in images, highlighting how language influences both expanding and biasing vision tasks.

\citet{fairrag} introduced FairRAG, focusing on mitigating biases in human generation tasks by incorporating fair retrieval methods. Their approach adjusts retrieval processes to include diverse and representative examples, reducing biases in generated content. On the other hand, \citet{finetuning} proposed fine-tuning text-to-image diffusion models for fairness using a single textual inversion token, demonstrating that targeted fine-tuning can mitigate demographic biases, although it may be computationally intensive and limited to specific models. Complementary to guidance- and token-based methods, \citet{huang2025ddm} propose a Debiasing Diffusion Model (DDM) that integrates a latent representation learning indicator into Stable Diffusion to promote fairer generations without predefined sensitive-attribute conditions, and to reduce group disparities even when attributes are specified. 

\paragraph{Language-Driven Approaches in Bias Mitigation}

Language-driven strategies have been proposed for bias mitigation. \citet{5} used text embedding projection for debiasing vision-language discriminative and generative models. On the other hand, \citet{12} proposed dataset debiasing by enhancing datasets with synthetic, gender-balanced sets. In text-to-image models, \citet{biasp1} suggested adding gender terms to prompts to balance gender representation in images. Similarly, \citet{biasp2} advocated adding ethical statements to prompts to directly encourage fairness.

\paragraph{Safety Benchmarks for T2I Models}

Beyond mitigation methods, standardized evaluation is crucial. \citet{li2025t2isafety} introduce \emph{T2ISafety}, a large-scale benchmark spanning three domains—fairness, toxicity, and privacy—with a hierarchy of 12 tasks and 44 categories, 70K prompts, and 68K human-annotated images. They also propose normalized KL divergence as a fairness metric and release an MLLM-based evaluator (ImageGuard), revealing persistent racial fairness issues and model-to-model variability across safety dimensions. 

\paragraph{Advances in Large Language Models}

Recent advances in LLMs have enhanced model reasoning and debiasing capabilities. \citet{sun2023enhancing} introduced iterative bootstrapping in Chain-of-Thought prompting to improve problem-solving capabilities in models like ChatGPT. \citet{star} proposed Self-Taught Reasoner (STaR), enabling models to self-supervise and refine reasoning steps without additional labeled data. Furthermore, \citet{faith} focused on improving the faithfulness of reasoning steps in LLMs, addressing hallucination, and enhancing reliability in sensitive contexts. However, \citet{biascot} highlighted risks of zero-shot CoT reasoning in socially sensitive domains, showing that models may generate biased or harmful content without proper guidance, underscoring the need for mechanisms to minimize biased outputs.

\paragraph{Attribute Prediction}

Attribute prediction is crucial for assessing demographic representations in generated content. CLIP \citep{clip} has been widely used for zero-shot image classification and attribute prediction due to its ability to learn visual concepts from natural language supervision. \citet{clip} found high accuracy (96\%) using CLIP for gender classification across all race class labels, with around 93\% for racial classification and 63\% for age classification. In addition, \citet{finetuning} adopted the eight race categories from the FairFace dataset but found classifiers struggled to distinguish between certain categories. Therefore, to improve race attribute prediction, they consolidated them into four broader classes: WMELH (White, Middle Eastern, Latino Hispanic), Asian (East Asian, Southeast Asian), Black, and Indian. \citet{han} presented a deep multi-task learning approach for heterogeneous face attribute estimation, emphasizing multi-task learning to improve attribute prediction accuracy and robustness across diverse populations.

Our work builds upon these studies by integrating \emph{iterative Chain-of-Thought reasoning} within multimodal generative LLMs to enhance fairness in T2I models. Instead of relying on labor-intensive prompt engineering or costly model fine-tuning, our approach dynamically refines textual prompts through real-time feedback, effectively avoiding the pitfalls of oversimplified zero-shot CoT in socially sensitive settings. Additionally, an \emph{attire-based} detection mechanism captures subtle markers—such as religious attire—to broaden the scope of bias identification beyond standard demographic attributes. Notably, our framework generalizes to scenarios beyond single-person or profession-centric content (e.g., children, animals, objects), promoting more inclusive, ethically responsible T2I generation across diverse use cases.

\section{Methods}
\label{sec:methodology}

In this section, we introduce \textbf{FairCoT}, a framework that integrates MLLMs to enhance fairness in AI-generated imagery. We specifically leverage the \emph{reasoning} capabilities of MLLMs to improve fairness in T2I models that otherwise lack detailed reasoning or have limited input tokens. FairCoT operates in two key phases: a \emph{CoT Generation Phase} and an \emph{Inference Phase}.

\subsection{Chain-of-Thought Generation Phase}

\paragraph{Initial Image Generation.}

We begin by defining a set of professions \(\mathcal{P} = \{p_1, p_2, \dots, p_n\}\) that span a broad range of societal roles, encompassing Healthcare and Medical, Legal and Business, Service and Hospitality, Security and Protection, Education and Information, Engineering and Technical, and Research and Analytics. We also identify demographic attributes \(\mathcal{D} = \{\text{gender}, \text{age}, \text{race}, \text{religion}\}\). For religion, we specify a set of groups \(\mathcal{R} = \{r_1, r_2, \dots, r_k\}\).

A Multimodal LLM generates initial prompts \(\mathcal{S} = \{s_1, s_2, \dots, s_n\}\) of the form “n photos of \(p_i\).” These prompts guide a T2I diffusion model to produce an initial set of images \(\mathcal{I} = \{I_1, I_2, \dots, I_m\}\). This initial generation step provides raw data for subsequent bias assessments and refinements.

\paragraph{Attribute Prediction}

Next, we apply CLIP for zero-shot classification of key attributes: \emph{gender}, \emph{race}, \emph{age}, and \emph{religion}. Since some attributes (e.g., religious attire) may be challenging for CLIP to identify directly, we augment its predictions with an \emph{attire-based} method. Specifically, we prompt the LLM to list attires \(\mathcal{A} = \{a_1, a_2, \dots, a_l\}\) that correspond to each religious group \(r_j \in \mathcal{R}\) (e.g., hijabs for Islam, turbans for Sikhism). 

For each image \(I_i \in \mathcal{I}\), we compute the cosine similarity:
\[
\text{Score}(I_i, a_j) = \cos\!\bigl(\phi_I(I_i), \phi_T(a_j)\bigr),
\]
where \(\phi_I\) and \(\phi_T\) are CLIP’s image and text encoders, respectively. The attire \(a_j\) yielding the highest similarity is used to infer the religious attribute for that image. This two-step process provides more robust and fine-grained predictions for demographic attributes.

Table~\ref{tab:agreement_comparison} contrasts the original (\emph{Vanilla}) CLIP predictions with our \emph{Enhanced} predictor, evaluated against hand-labeled (\emph{Hand}) data. By leveraging an attire-based LLM-driven strategy, the enhanced approach achieves a 75\% agreement rate with hand labels, substantially outperforming the 41.12\% achieved by the vanilla model. This improvement enables FairCoT to better identify and represent diverse religious affiliations in generated images.Further details are included  in App. \ref{sec:agreement}.We also benchmark CLIP against DeepFace on the same dataset for age, gender, and race, observing a better agreement (see \ref{app:agreement}).

\begin{table}[h]
\centering
\setlength{\tabcolsep}{4pt}
\small
\caption{Agreement with Hand-Labeled Religious Attributes}
\label{tab:agreement_comparison}
\begin{tabular}{@{}lcc@{}}
\toprule
Method & Agreement (\%) \\ \hline 
Vanilla Prediction & 41.12 \\
Enhanced Prediction (Ours) & \textbf{75.00} \\ 
\bottomrule
\end{tabular}
\end{table}

\begin{figure}[ht]
\centering
\includegraphics[width=0.4\textwidth]{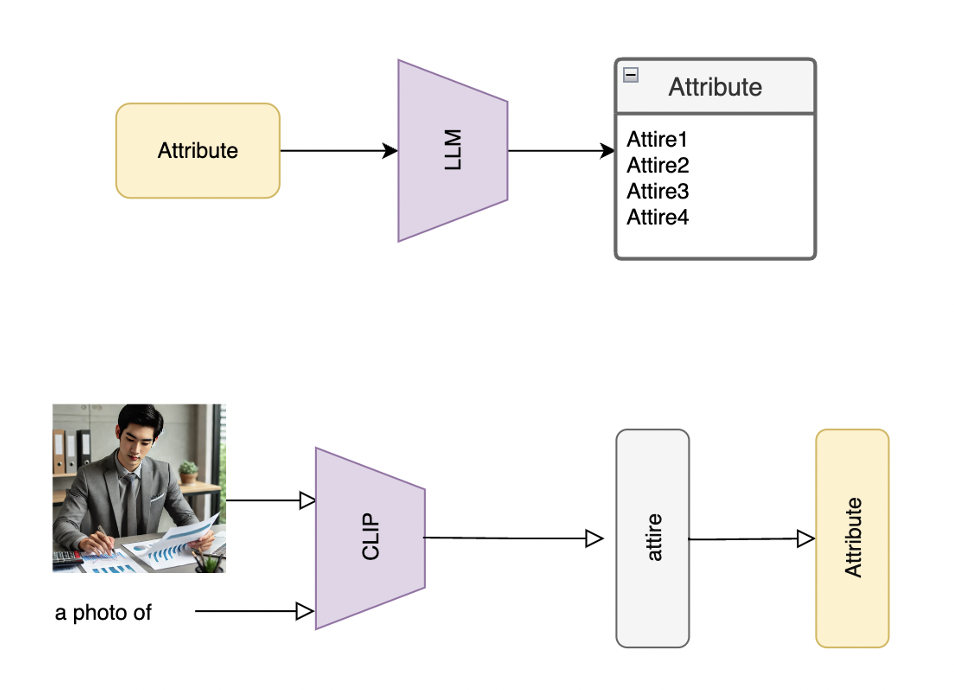} % Adjust width as necessary
\caption{Improving CLIP attribute prediction through an attire-based approach.}
\label{fig:img1}
\end{figure}

\paragraph{Bias and Alignment Evaluation \& Iterative Bootstrapping.}

We evaluate the distribution of demographic attributes in \(\mathcal{I}\) by computing the \emph{normalized entropy}:
\[
H' = -\left(\frac{1}{\log k} \sum_{j=1}^{k} p(a_j) \log p(a_j)\right),
\]
where \(p(a_j)\) is the empirical probability of attribute \(a_j\) (e.g., a specific religion) appearing in the image set, and \(k\) is the number of possible categories for that attribute. Note that \(H'\) ranges from 0 to 1, where 1 indicates perfectly even (i.e., fair) distribution across all categories.This even distribution assumption aligns with prior work in fair generation, where uniformity is commonly adopted as the target distribution \cite{faird,finetuning,5}. Normalized entropy thus serves as a principled measure of representational parity. Nevertheless, FairCoT is adaptable and can incorporate alternative target distributions when uniformity is not contextually appropriate.

To measure how well the generated images align with the original profession prompts, we calculate the \emph{CLIP-Textual} (CLIP-T) score:
\[
\text{CLIP-T} = \frac{1}{m} \sum_{i=1}^{m} \cos\!\bigl(\phi_I(I_i), \phi_T(s_i)\bigr),
\]
where \(s_i\) is the prompt used to generate the image \(I_i\). Higher scores indicate stronger semantic alignment with the original profession prompt (e.g., “a photo of a doctor”). CLIP-T ranges from \(-1\) to \(1\), so while some methods may enhance fairness (e.g., showing more females), they could inadvertently decrease alignment (e.g., generating a female image but not of a doctor).

\paragraph{CoT Debiasing.}

We initialize the model with a baseline Chain-of-Thought (\(\text{CoT}_0\)) from the zero-shot debiasing approach in \citet{think}, employing an instruction such as: 

\emph{“Think step by step before generating images while considering several races, genders, religions, and ages, and treating people of these categories equally.”}At each iteration \(t\):
\begin{enumerate}
    \item \textbf{Generate Images:} Use the current \(\text{CoT}_t\) to prompt the diffusion model, producing a new set of images \(\mathcal{I}_t\).
    \item \textbf{Evaluate Bias and Alignment:} Compute \(H'_t\) and the CLIP-T score for \(\mathcal{I}_t\).
    \item \textbf{Refine CoT:} If \(H'_t\) suggests persistent imbalance and CLIP-T remains sufficiently high (above a threshold \(\tau \cdot \text{CLIP-T}_{t_0}\)), then we update \(\text{CoT}_t\). We prompt the MLLM to “think again,” focusing specifically on improving fairness (e.g., “Can you think again? Consider generating images of different religions, races, ages, and genders.”).
    \item \textbf{Convergence Check:} If fairness (\(H'_t\)) does not improve compared to \(H'_{t-1}\) or if alignment (CLIP-T) falls below \(\tau \cdot \text{CLIP-T}_{t_0}\), we stop updating \(\text{CoT}\).
\end{enumerate}

\begin{algorithm}[t]
\small
\caption{FairCoT Iterative Refinement Process}
\label{alg:faircot}
\KwIn{Initial CoT (\(\text{CoT}_{t_0}\)), initial bias metric (\(H'_{t_0}\)), threshold \(\tau\).}
\KwOut{Refined CoT and fair images \(\mathcal{I}_t\).}
\For{\(t = t_0, t_0+1, \ldots\)}{
\textbf{Image Generation:} Generate \(\mathcal{I}_t\) using \(\text{CoT}_t\).\\
\textbf{Bias \& Alignment Evaluation:} Compute \(H'_t\) and \(\text{CLIP-T}_t\).\\
\textbf{Refinement:} \\
\If{(\(H'_t > H'_{t-1}\)) AND \(\text{CLIP-T}_t > \tau \cdot \text{CLIP-T}_{t_0}\)}{
    \textbf{Update} \(\text{CoT}_t\) (prompt MLLM with ``Can you think again? Consider generating images of different religions, races, ages, and genders.'')
}
\Else{
    \textbf{Stop} (\text{CoT} remains the same).
}
}
\end{algorithm}

\begin{figure*}[t!]
\begin{center}
\includegraphics[width=4.5in]{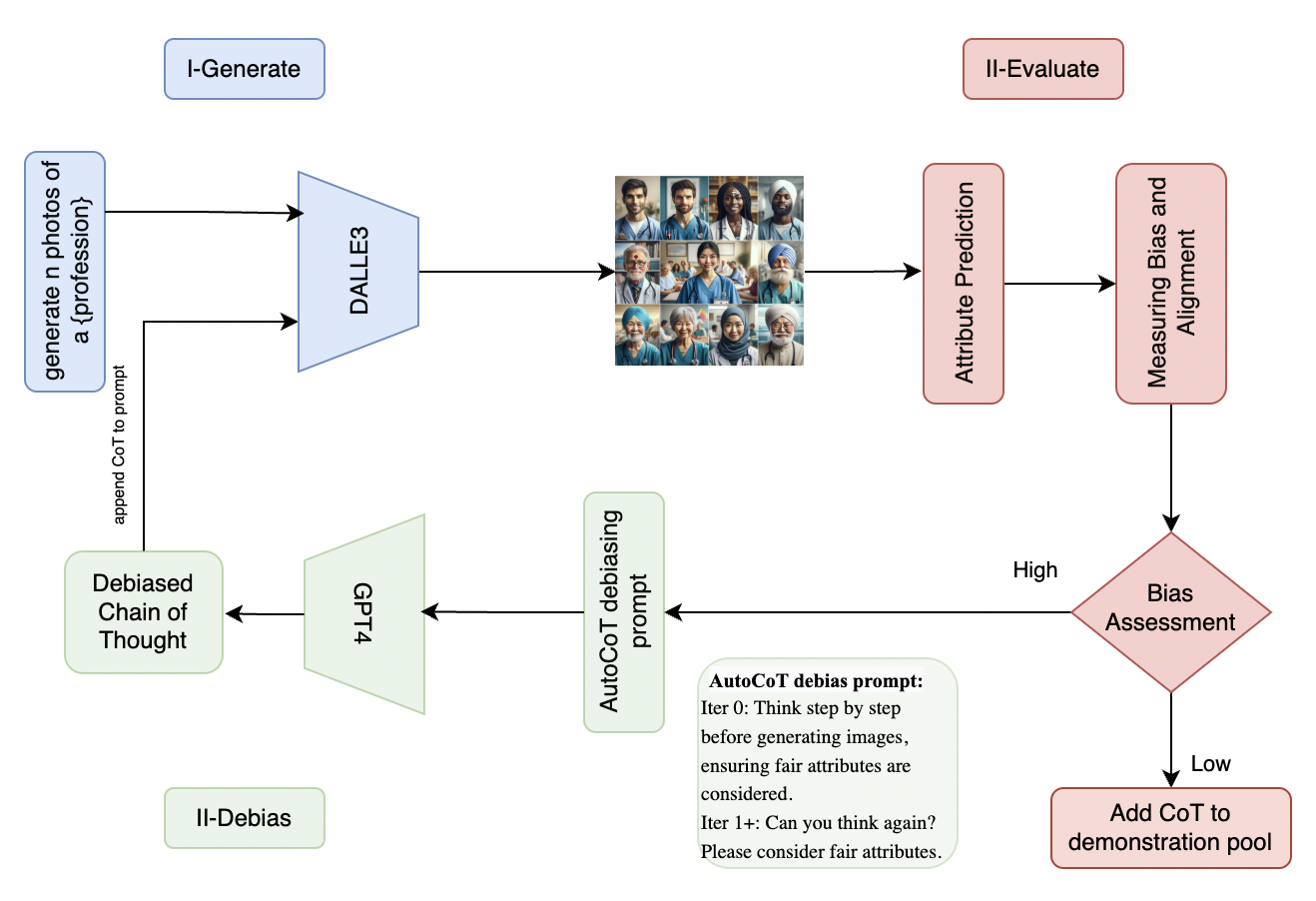}
\end{center}
\caption{CoT Generation: The process involves iterative steps of image generation, followed by bias and alignment evaluation, and then potential debiasing via an updated Chain-of-Thought. We use an MLLM (e.g., GPT-4 with DALLE) capable of reasoning and image generation.}
\label{fig:img2}
\end{figure*}

Figure~\ref{fig:img2} illustrates the overall workflow, showcasing how image generation, bias evaluation, and CoT refinement are repeated until the fairness criteria converge (i.e., when \(H'_t\) no longer improves or alignment drops below the acceptable threshold).

\subsection{Inference Phase}

\paragraph{Demonstration Pool and Application.}

Once the generated images and corresponding \(\text{CoT}\) reasoning achieve satisfactory fairness, we archive them in a \emph{demonstration pool} \(\mathcal{D}_{\text{pool}}\). This repository records the final Chain-of-Thought, prompts, and profession labels, serving as a reusable template for future tasks.

\paragraph{Inference.}

For a new task or a new profession \(p_{\text{new}}\), we select an appropriate \(\text{CoT}\) from \(\mathcal{D}_{\text{pool}}\) based on its similarity to the targeted domain. The MLLM adapts this archived \(\text{CoT}\) to generate a task-specific \(\text{CoT}_{\text{new}}\). It then produces prompts \(\mathcal{S}_{\text{new}}\) that incorporate fairness considerations, guiding the T2I to generate images \(\mathcal{I}_{\text{new}}\). This approach preserves learned fairness criteria and easily generalizes beyond professions or single-person generations without additional retraining or parameter modifications. Figure~\ref{fig:img1} illustrates how FairCoT’s debiasing extends to new tasks and domains.

\begin{figure*}[t!]
\begin{center}
\includegraphics[scale=0.5]{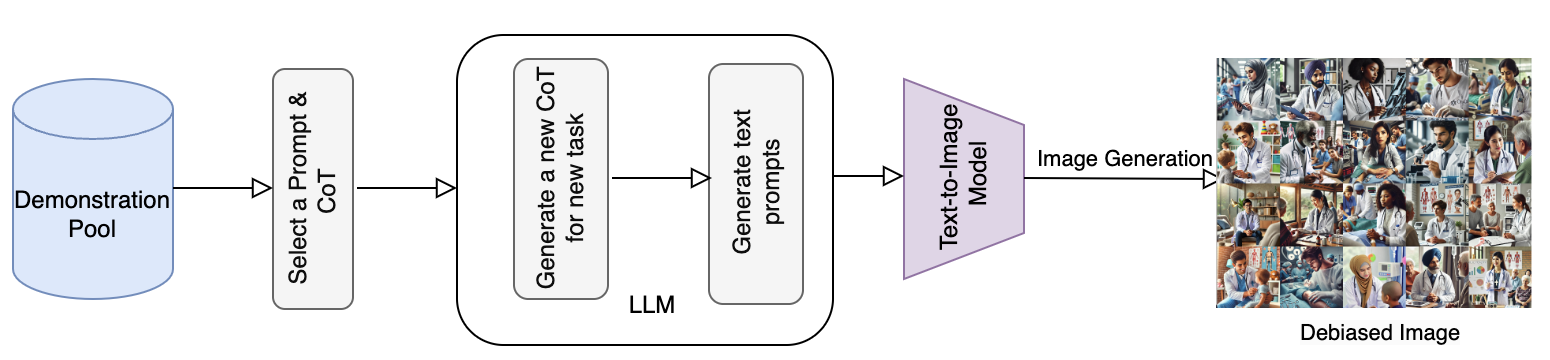} % Replace with the path to your image
\end{center}
\caption{FairCoT Inference: An MLLM uses archived Chain-of-Thought demonstrations to produce fair prompts for T2I models, ensuring consistent debiasing on novel tasks.}
\label{fig:img1}
\end{figure*}

\section{Experimental Results}

We conducted a comprehensive set of experiments to evaluate the effectiveness of \textbf{FairCoT} in enhancing fairness and diversity in T2I diffusion models. Inspired by the methodologies of \citet{sun2023enhancing, star, biascot, finetuning}, our experiments were designed to assess FairCoT’s bias-mitigation capabilities while preserving semantic alignment. 
\subsection{Experimental Setup}
\subsubsection{CoT Generation \& Inference Professions}

As detailed in \ref{sec:professions}, we used specific professions during the CoT generation (training) phase and tested on related but distinct professions during the inference phase. For instance, in the \emph{Healthcare and Medical Professions} domain, we employed \emph{Nurse} as a CoT generation profession and evaluated the model on \emph{Doctor}, \emph{Pharmacist}, and \emph{Dentist} during inference. This strategy was replicated across various professional domains  to ensure the robustness of our approach. Refer to \ref{sec:professions} for the complete list of CoT-gen and test professions.
\subsubsection{Models Evaluated}
\label{sec:models_evaluated}

We assessed two categories of T2I models priority and open-source. In the priority category, we used DALL-E for its high-fidelity image outputs added to reasoning capability through GPT4. In the open-source category, our evaluation focused on three variants of Stable Diffusion that can't reason: SDv1-5\citep{sdv2}, which serves as a baseline version; SDv2-1 \citep{sdv2}, an updated release with enhanced capabilities; and SDXL-turbo \citep{sdxl}, optimized for real-time performance. 

All T2I models were tested under a variety of methods. General Prompting employed standard prompts without any bias mitigation. Ethical Intervention \citep{biasp2} involved augmenting prompts with ethically oriented statements. DebiasVL \citep{5} applied a debiasing technique based on text embedding projection, while FairD \citep{faird} used a diffusion-based debiasing method. We also tested Finetune \citep{finetuning}, in which models are fine-tuned specifically for fairness. Finally, FairCoT (Ours) introduced our Chain-of-Thought approach, evaluated in both full-body and headshot settings.

In addition, we experimented with different MLLMs to generate the Chain-of-Thought for \textbf{FairCoT} at inference time. Specifically, we tested GPT-4 and LLaMA V3.2 11B Instruct to demonstrate the flexibility of our framework when adapting to various MLLMs.

Four demographic attributes were considered: \emph{gender}, \emph{race}, \emph{age}, and \emph{religion}. Following \citet{finetuning}, gender was categorized as female or male; race included WMELH, Asian, Black, and Indian; age was split into young or old; and religion encompassed Islam, Christianity, Hinduism, and a \emph{neutral} category for images without explicit religious attire or symbols \citep{cia_religions_2023}.

\subsubsection{Evaluation Metrics}
\label{sec:evaluation_metrics}

We measure fairness and diversity using the \emph{Bias-Normalized Entropy} (\(H'\)), which ranges from 0 to 1. Higher values indicate greater uniformity of attribute distributions (e.g., gender, race, age, religion) in generated images. To measure semantic alignment between generated images and the original prompts, we compute the CLIP-T score. This metric ensures that gains in fairness do not compromise semantic relevance.

\subsection{Results}
\label{sec:experiments}

\subsubsection{General Results}

Table~\ref{tab:experimental_results} presents the \emph{Bias-Normalized Entropy} (\(H'\)) and \emph{CLIP-T} scores for 20 professions under different models and debiasing strategies. Across all settings, \textbf{FairCoT} achieves the highest \(H'\) scores for gender, race, age, and religion in most comparisons, outperforming baseline methods by substantial margins. For instance, under SDv1-5, the gender entropy improves from 0.47 (General) to 0.97 (FairCoT), while race entropy increases from 0.55 (General) to 0.96 (FairCoT). Meanwhile, \emph{CLIP-T} scores remain competitive, signifying minimal impact on semantic coherence. App. \ref{app:cmmd_kid} on image quality analysis shows FairCoT attains the best CMMD (0.023) and KID (0.069) scores, and \ref{app:compute} reports a 97 s inference for 20 images—≈17 s slower than a generic prompt yet still faster than FairD or DebiasVL—confirming both quality and efficiency.

\begin{table}[h]
\centering
\setlength{\tabcolsep}{4pt}
\small
\caption{Comparison of FairCoT with DebiasVL \citep{5}, Ethical Intervention \citep{biasp2}, Finetune \citep{finetuning}, and FairD \citep{faird} across 20 professions. \emph{G}: Gender, \emph{R}: Race, \emph{A}: Age, \emph{Rl}: Religion, \emph{CT}: CLIP-T.}
\label{tab:experimental_results}
\resizebox{\columnwidth}{!}{
\begin{tabular}{@{}llccccr@{}}
\toprule
& & \multicolumn{4}{c}{Bias-Normalized Entropy$\uparrow$} & \multicolumn{1}{c}{Gen} \\ 

Model & Prompt & G & R & A & Rl& CT$\uparrow$ \\ \hline

\multirow{3}{*}{DALLE-gen} & General & 0.56  & 0.38  & 0.68  & 0.33  & \textbf{0.27}   \\
& Ethical Int. & 0.84  & 0.67  & 0.7   & 0.36  & \textbf{0.27}  \\
& Ours & \textbf{0.93}  & \textbf{0.83}  & \textbf{0.9}   & \textbf{0.68}  & 0.26 \\ \hline

\multirow{3}{*}{DALLE-test} & General & \textbf{0.99} & 0.89& 0.23& 0.27& \textbf{0.27}
  \\
& Ethical Int. & 0.87  & 0.65  & 0.29   & 0.59  & 0.26 \\
& Ours & \textbf{0.99}  & \textbf{0.95}  & \textbf{0.57}   &\textbf{0.75}  & 0.26 \\ \hline

\multirow{8}{*}{SDv1-5} & General &  0.47  & 0.55  & 0.28  & 0.27  & \textbf{0.28} \\
& Ethical Int.& 0.72  & 0.56  & 0.27  & 0.32  & 0.27 \\
& FairD. & 0.31  & 0.52  & 0.17  & 0.39  & 0.26  \\
& DebiasVL & 0.08  & 0.23  & \textbf{0.61}  & 0.22  & 0.27 \\
& Finetune & 0.96  & 0.74  & 0.25  & 0.28  & 0.26 \\
& Ours & 0.97  & \textbf{0.96}  & 0.49  & \textbf{0.85}  & 0.26 \\
& Ours-face & \textbf{0.98}  & \textbf{0.96}  & 0.47  & 0.78  & 0.25 \\ 
& Ours-llama & 0.92	&0.85	&0.43	&0.74	&0.25\\\hline

\multirow{5}{*}{SDXL-turbo} & General & 0.25  & 0.38  & 0.35  & 0.24  & \textbf{0.30}  \\
& Ethical Int. & 0.33  & 0.08  & 0.16  & 0.22  & 0.28 \\
& Ours & 0.98  & 0.92  & 0.43  & 0.84  & 0.26 \\
& Ours-face & \textbf{0.99}  & \textbf{0.94}  & \textbf{0.60}  & \textbf{0.92}  & 0.26  \\ 
& Ours-llama & 0.93&	0.77&	0.41&	0.77	&0.26\\ \hline

\multirow{6}{*}{SDV2-1} & General & 0.50  & 0.55  & 0.40  & 0.36  & 0.28 \\
& Ethical Int. & 0.58  & 0.46  & 0.20  & 0.38  & 0.26  \\
& DebiasVL & 0.56  & 0.50  & 0.26  & 0.25  & \textbf{0.29} \\
& Ours & \textbf{0.98}  & 0.95  & 0.38  & 0.85  & 0.26 \\
& Ours-face & \textbf{0.98}  & \textbf{0.96}  & \textbf{0.43}  & \textbf{0.87}  & 0.26 \\& Ours-llama & 	0.92&	0.86&	0.39&	0.72&	0.25 \\
\bottomrule
\end{tabular}
}
\end{table}

\subsubsection{Generalization to Multiface Generation}
\label{sec:multiface_results}

We also examined \textbf{FairCoT} in a \emph{multiface} setting, where each image contains three faces. Table~\ref{tab:multiface} shows results over 10 professions. Even in this more complex scenario, FairCoT continues to provide high diversity scores. Under SDv1-5, for example, the gender entropy increases from 0.57 (General) to 0.95 (FairCoT), while the race entropy improves from 0.47 to 0.84. These gains illustrate FairCoT’s capacity to handle multiple faces with minimal impact on semantic coherence.Implementation details appear in  \ref{app:multiface_details}.

\begin{table}[h]
\centering
\setlength{\tabcolsep}{4pt}
\small
\caption{Evaluation in a Multi-Face Setting (three faces per image)}
\label{tab:multiface}
\resizebox{\columnwidth}{!}{%
\begin{tabular}{@{}llccccr@{}}
\toprule

& & \multicolumn{4}{c}{Bias-Normalized Entropy$\uparrow$} & \multicolumn{1}{c}{Gen} \\ 

Model & Prompt & G & R& A & Rl &CT$\uparrow$ \\ \hline

\multirow{3}{*}{DALL-E} & General & 0.76  & 0.76  & \textbf{0.89}  & 0.71  & 0.23 \\
& Ethical Int. & 0.94  & 0.83  & 0.63  & 0.66  & \textbf{0.26} \\
& Ours & \textbf{0.95}  & \textbf{0.88}  & 0.70  & \textbf{0.82}  & \textbf{0.26}  \\ \hline

\multirow{5}{*}{SDv1-5} & General & 0.57  & 0.47  & 0.41  & 0.36  & \textbf{0.27}  \\
& Ethical Int. & 0.77  & 0.63  & 0.23  & 0.44  & 0.26  \\
& Finetune & \textbf{0.96}  & 0.71  & 0.27  & 0.50  & \textbf{0.27}  \\
& Ours & 0.95  & 0.84  & \textbf{0.51}  & \textbf{0.81}  & 0.25 \\
& Ours-llama & 0.87  & \textbf{0.86}  & 0.44  & 0.76  & 0.25 \\\hline

\multirow{4}{*}{SDXL-turbo} & General & 0.43  & 0.31  & 0.39  & 0.36  & \textbf{0.27}  \\
& Ethical Int. & 0.65  & 0.48  & 0.31  & 0.40  & 0.24  \\
& Ours & \textbf{0.82}  & \textbf{0.80}  & \textbf{0.59}  & \textbf{0.79}  & 0.24  \\
& Ours-llama& 0.80  & \textbf{0.80}  & 0.47  & 0.60  & 0.25 \\\hline

\multirow{4}{*}{SDv2-1} & General & 0.58  & 0.40  & 0.49  & 0.46  & \textbf{0.26} \\
& Ethical Int. & 0.75  & 0.59  & 0.24  & 0.66  & 0.22  \\
& Ours  & 0.87  & 0.79  & 0.40  & \textbf{0.76}  & 0.25 \\
& Ours-llama & \textbf{0.88}  & \textbf{0.84}  & \textbf{0.52}  & 0.74  & 0.25\\
\bottomrule
\end{tabular}
}
\end{table}

\subsubsection{Generalization to Multiconcept Generation}
\label{sec:multiconcept_results}

We further evaluated \textbf{FairCoT} in \emph{multiconcept} scenarios, where prompts combine multiple entities—for example, adults, children, animals, or objects such as laptops and dog breeds. As shown in Table~\ref{tab:multiconcept}, FairCoT achieves near-uniform attribute distributions even when handling multiple concepts simultaneously. Under SDv1-5, for instance, the gender entropy jumps from 0.27 (General) to 0.99 (FairCoT), while race entropy improves from 0.61 to 0.93.  Moreover, our method overcomes the Apple computer bias in DALL-E \citep{mac} by generating a diverse range of computer brands—Apple, Microsoft, Asus, HP, and Dell. These results highlight FairCoT’s robust generalization to more complex generative tasks without sacrificing overall quality or alignment (further details in \ref{appendix:qualitative_results}).

\begin{table}[h]
\centering
\setlength{\tabcolsep}{4pt}
\small
\caption{Evaluation in Multi-Concept Generation}
\label{tab:multiconcept}
\resizebox{\columnwidth}{!}{
\begin{tabular}{@{}llccccr@{}}
\toprule
& & \multicolumn{4}{c}{Bias-Normalized Entropy$\uparrow$} & \multicolumn{1}{c}{Gen} \\ 

Model & Prompt & G & R & A & Rl & CT$\uparrow$ \\ \hline

\multirow{3}{*}{DALL-E} & General & 0.69  & 0.68  & \textbf{0.65}  & 0.25  & \textbf{0.28} \\
& Ethical Int. & 0.77  & 0.71  & 0.35  & 0.58  & 0.26  \\
& Ours & \textbf{0.97}  & \textbf{0.93}  & \textbf{0.65}  & \textbf{0.73}  & \textbf{0.28} \\ \hline

\multirow{5}{*}{SDv1-5} & General & 0.27  & 0.61  & 0.50  & 0.47  & \textbf{0.30} \\
& Ethical Int. & 0.56  & 0     & 0     & 0.24  & \textbf{0.30}  \\
& Finetune & 0.81  & 0.82  & 0.58  & 0.45  & 0.29\\
& Ours  & \textbf{0.99}  & 0.93  & 0.39  & \textbf{0.79}  & 0.27 \\
& Ours-llama & 0.97  & \textbf{0.95}  & \textbf{0.69}  & 0.72  & 0.27 \\\hline

\multirow{4}{*}{SDXL-turbo}& General & 0.43  & 0.54  & 0.33  & 0.35  & \textbf{0.30}   \\
& Ethical Int.& 0.34  & 0     & 0     & 0.29  & 0.29  \\
& Ours &  \textbf{0.98}  & \textbf{0.88}  & 0.40  & \textbf{0.86}  & 0.27  \\
& Ours-llama & 0.90  & 0.84  & \textbf{0.58}  & 0.68  & 0.27 \\\hline

\multirow{4}{*}{SDv2-1} & General& 0.79  & 0.66  & 0.53  & 0.50  & \textbf{0.29}  \\
& Ethical Int.&  0.72  & 0.55  & 0.27  & 0.60  & 0.28  \\
& Ours& \textbf{0.99}  & \textbf{0.87}  & 0.71  & \textbf{0.85}  & 0.26  \\
&Ours-llama & 0.95  & 0.84  & \textbf{0.74}  & 0.63  & 0.27 \\
\bottomrule
\end{tabular}
}
\end{table}

Finally, we conducted an ablation study to assess the impact of each FairCoT component on performance. Details and discussions are presented in Appendix~A. Overall, these results emphasize FairCoT’s robust generalization to various generative contexts—ranging from single- to multi-face scenarios and simple to multi-concept prompts—while maintaining strong semantic alignment.

\subsection{Discussion}
\label{sec:discussion}

Our experimental findings demonstrate that \textbf{FairCoT} significantly boosts fairness and diversity in T2I diffusion models across a wide spectrum of scenarios and architectures. By achieving consistently higher \emph{Bias-Normalized Entropy} scores, FairCoT effectively reduces demographic imbalances without substantially compromising image quality, as evidenced by stable or only slightly reduced \emph{CLIP-T} scores. This performance balance underscores FairCoT’s capacity to address multi-attribute biases while maintaining semantic coherence.

Compared with fine-tuning strategies \citep{finetuning}, FairCoT requires neither model retraining nor intrusive parameter updates, making it flexible and model-agnostic. Fine-tuning methods can be computationally demanding, expensive, and less adaptable to diverse T2I architectures or prompt variations. By contrast, FairCoT introduces fairness interventions at the \emph{reasoning level}, leveraging the iterative \emph{CoT} capabilities of MLLMs. This not only streamlines deployment in both closed-source (e.g., DALL-E) and open-source (e.g., Stable Diffusion) models but also generalizes beyond traditional profession-oriented prompts. Indeed, our approach readily extends to multiple domains—ranging from generating images of children, animals, and household objects to multi-faced scenes—illustrating its robustness and adaptability to various generative tasks.

Qualitative examples in the appendix further highlight FairCoT’s effectiveness in generating demographically diverse outputs. In many baseline models, real-world biases ingrained in the training data surface as stereotypical or underrepresented depictions of certain groups. FairCoT alleviates these concerns through iterative CoT refinements, where explicit fairness constraints are embedded into the prompt-generation phase. Rather than relying solely on static prompt manipulation, our framework incorporates evolving ethical guidance, thus achieving broader coverage of sensitive demographic categories.

Despite these advantages, there remain avenues for future exploration. First, FairCoT depends on \emph{language-driven reasoning} to guide multimodal generation, which could still inherit biases from the underlying MLLM. Second, while iterative CoT procedures reduce biases across a range of common attributes (e.g., race, gender, age, religion), further work is needed to tackle more complex or context-specific biases—such as intersectional biases involving geographic or cultural domains. Lastly, integrating FairCoT with complementary \emph{post hoc} debiasing tools or human-in-the-loop systems could refine the fairness criteria even further. Overall, our results indicate that reasoning-level interventions, when paired with robust prompt generation, can substantially advance the goal of fair and inclusive AI-driven image generation.

\section{Conclusion}
\label{sec:conclusion}

We have presented \textbf{FairCoT}, a Chain-of-Thought-based framework that harnesses iterative, MLLM-powered reasoning to mitigate biases in text-to-image diffusion models without compromising semantic alignment. By transparently guiding the generative process, FairCoT effectively addresses demographic imbalances across attributes such as gender, race, age, and religion. Our experiments demonstrate that FairCoT outperforms both prompt-engineering and fine-tuning approaches, underscoring its model-agnostic applicability across priority and open-source systems.

Given FairCoT’s versatility, we anticipate its adoption in a broad range of domains, from creative content generation to large-scale enterprise applications. Looking ahead, our research will explore more context-specific biases, investigate deeper integration with user feedback loops, and refine the iterative Chain-of-Thought mechanism to continuously incorporate evolving ethical considerations. We believe this work offers a practical and principled roadmap for embedding fairness-driven strategies within multimodal AI systems, fostering a new generation of responsible, inclusive, and socially aware technologies.

\section*{Acknowledgements}
MP was supported by the UK Engineering and Physical Sciences Research Council (grant number EP/Y009800/1) through funding from Responsible AI UK (project KP0016, “AdSoLve: Addressing Socio-technical Limitations of LLMs for Medical and Social Computing”), and by the Slovenian Research and Innovation Agency (ARIS) through the Gravitacije project LLM4DH (“Large Language Models for Digital Humanities”, GC-0002), the project CroDeCo (“Cross-Lingual Analysis for Detection of Cognitive Impairment in Less-Resourced Languages”, J6-60109) and the research programme “Knowledge Technologies” (P2-0103). This work made use of the Apocrita HPC facility at Queen Mary University of London (QMUL), supported by the QMUL Research-IT team, and benefited from compute provided through the OpenAI Researcher Access Program. ZA is supported by Google DeepMind PhD Fellowship and thanks their Google DeepMind mentor, David Stutz, for guidance and support..

\section*{Limitations}
While FairCoT shows substantial improvements, certain limitations exist. First, our approach uses attribute classifiers to measure bias and diversity. This relies on accurate attribute predictions, which can vary in reliability across different demographic groups and attribute types—particularly those that are culturally or contextually sensitive. Future research could investigate incorporating more robust, domain-specific, or ensemble-based attribute prediction models to reduce the risk of misclassification.

Second, although we address religious bias by diversifying attire, we do not imply that clothing choices perfectly represent religious identity. Attire often serves as an overt signifier of cultural and religious practices, but it does not capture the full spectrum of religious expression or personal beliefs. This limitation underscores the need for more advanced interventions—incorporating broader cultural context, self-identification, and intersectional attributes—to avoid simplifying or misrepresenting religious identities.

Additionally, our focus has been on a set of commonly studied attributes (e.g., race, gender, religion). Extending FairCoT to biases related to disability, body types, socio-economic status, or intersectional categories (e.g., disabled women from specific cultural backgrounds) remains an open challenge. Many of these attributes are difficult to detect visually or may intersect in ways that compound bias in generated content. Employing \emph{multimodal chain-of-thought} approaches, where linguistic and visual cues jointly inform bias detection and mitigation, could further enhance fairness across diverse domains.

Further, FairCoT operates without modifying training data or model parameters, so its ability to elicit “fair” generations is bounded by what the underlying models have learned. For attributes or criteria that occur infrequently in the real world (e.g., rare occupations, specific cultural practices, certain disabilities, or low-prevalence intersections), base models may have limited exposure. In such long-tail settings, even when FairCoT prompts for coverage, there is a risk of semantic degradation—e.g., reduced fidelity, implausible co-occurrences, or attenuated depiction quality. We did not directly quantify this effect, but we acknowledge it as a plausible limitation. Future work should include targeted stress tests on low-occurrence phenomena, explicit reporting of coverage/failure rates, and mitigations such as retrieval-augmented prompting, small curated exemplars for rare attributes, or lightweight adapters to better support long-tail cases.

Moreover, FairCoT’s strategy of balancing attributes assumes there are scenarios where equal or near-equal representation is desirable. However, there may be cases where neither strictly equal nor proportional distributions are the intended goal—particularly if real-world prevalence or ethical considerations favor certain representations over others. Designing \emph{context-aware} fairness mechanisms that adapt to such cases while preserving authenticity is an important direction for future research.

Finally, FairCoT relies on MLLMs that can generate images in the CoT generation phase, and many of these models remain partially or fully closed-source. Ensuring transparent, open-sourced foundations for image generation is critical not only for reproducibility but also for allowing the broader research community to refine fairness interventions. Future work on open-sourcing advanced generative models—and integrating human-in-the-loop or \emph{post hoc} debiasing steps—can further enhance FairCoT’s applicability and foster a more inclusive ecosystem for AI-generated content.

% Bibliography entries for the entire Anthology, followed by custom entries
%\bibliography{anthology,custom}
% Custom bibliography entries only
\bibliography{custom.bib}

\begin{thebibliography}{26}
\providecommand{\natexlab}[1]{#1}

\bibitem[{Bansal et~al.(2022)Bansal, Yin, Monajatipoor, and Chang}]{biasp2}
Hritik Bansal, Da~Yin, Masoud Monajatipoor, and Kai-Wei Chang. 2022.
\newblock How well can text-to-image generative models understand ethical natural language interventions?
\newblock \emph{arXiv preprint arXiv:2210.15230}.

\bibitem[{Bianchi et~al.(2023)Bianchi, Kalluri, Durmus, Ladhak, Cheng, Nozza, Hashimoto, Jurafsky, Zou, and Caliskan}]{biasp1}
Federico Bianchi, Pratyusha Kalluri, Esin Durmus, Faisal Ladhak, Myra Cheng, Debora Nozza, Tatsunori Hashimoto, Dan Jurafsky, James Zou, and Aylin Caliskan. 2023.
\newblock Easily accessible text-to-image generation amplifies demographic stereotypes at large scale.
\newblock In \emph{Proceedings of the 2023 ACM Conference on Fairness, Accountability, and Transparency}, pages 1493--1504.

\bibitem[{Cate et~al.(2017)Cate, Dalvi, and Hussain}]{deepface}
Hardie Cate, Fahim Dalvi, and Zeshan Hussain. 2017.
\newblock Deepface: Face generation using deep learning.
\newblock \emph{arXiv preprint arXiv:1701.01876}.

\bibitem[{Chuang et~al.(2023)Chuang, Jampani, Li, Torralba, and Jegelka}]{5}
Ching-Yao Chuang, Varun Jampani, Yuanzhen Li, Antonio Torralba, and Stefanie Jegelka. 2023.
\newblock \href {https://arxiv.org/abs/arXiv:2302.00070v1} {Debiasing vision-language models via biased prompts}.

\bibitem[{{CIA}(2023)}]{cia_religions_2023}
{CIA}. 2023.
\newblock \href {https://www.cia.gov/the-world-factbook/field/religions/} {Religions - the world factbook}.
\newblock Retrieved 28 April 2023.

\bibitem[{D'Inc\`a et~al.(2024)D'Inc\`a, Peruzzo, Mancini, Xu, Goel, Xu, Wang, Shi, and Sebe}]{mac}
Moreno D'Inc\`a, Elia Peruzzo, Massimiliano Mancini, Dejia Xu, Vidit Goel, Xingqian Xu, Zhangyang Wang, Humphrey Shi, and Nicu Sebe. 2024.
\newblock Openbias: Open-set bias detection in text-to-image generative models.
\newblock In \emph{Proceedings of the IEEE/CVF Conference on Computer Vision and Pattern Recognition (CVPR)}, pages 12225--12235.

\bibitem[{Friedrich et~al.(2023)Friedrich, Brack, Struppek, Hintersdorf, Schramowski, Luccioni, and Kersting}]{faird}
Felix Friedrich, Manuel Brack, Lukas Struppek, Dominik Hintersdorf, Patrick Schramowski, Sasha Luccioni, and Kristian Kersting. 2023.
\newblock Fair diffusion: Instructing text-to-image generation models on fairness.
\newblock \emph{arXiv preprint arXiv:2302.10893}.

\bibitem[{Gandikota et~al.(2024)Gandikota, Orgad, Belinkov, Materzy{\'n}ska, and Bau}]{biasp3}
Rohit Gandikota, Hadas Orgad, Yonatan Belinkov, Joanna Materzy{\'n}ska, and David Bau. 2024.
\newblock Unified concept editing in diffusion models.
\newblock In \emph{Proceedings of the IEEE/CVF Winter Conference on Applications of Computer Vision}, pages 5111--5120.

\bibitem[{Hall et~al.(2023)Hall, Gustafson, Adcock, Misra, and Ross}]{6}
Melissa Hall, Laura Gustafson, Aaron Adcock, Ishan Misra, and Candace Ross. 2023.
\newblock \href {https://arxiv.org/abs/arXiv:2301.11100v1} {Vision-language models performing zero-shot tasks exhibit gender-based disparities}.

\bibitem[{Han et~al.(2017)Han, Jain, Wang, Shan, and Chen}]{han}
Hu~Han, Anil~K Jain, Fang Wang, Shiguang Shan, and Xilin Chen. 2017.
\newblock Heterogeneous face attribute estimation: A deep multi-task learning approach.
\newblock \emph{IEEE transactions on pattern analysis and machine intelligence}, 40(11):2597--2609.

\bibitem[{Huang et~al.(2025)Huang, Tsao, Su, and Chiang}]{huang2025ddm}
Lin-Chun Huang, Ching~Chieh Tsao, Fang-Yi Su, and Jung-Hsien Chiang. 2025.
\newblock \href {https://doi.org/10.48550/arXiv.2503.12536} {Debiasing diffusion model: Enhancing fairness through latent representation learning in stable diffusion model}.
\newblock \emph{arXiv preprint arXiv:2503.12536}.

\bibitem[{Jayasumana et~al.(2024)Jayasumana, Ramalingam, Veit, Glasner, Chakrabarti, and Kumar}]{cmmd}
Sadeep Jayasumana, Srikumar Ramalingam, Andreas Veit, Daniel Glasner, Ayan Chakrabarti, and Sanjiv Kumar. 2024.
\newblock Rethinking fid: Towards a better evaluation metric for image generation.
\newblock In \emph{Proceedings of the IEEE/CVF Conference on Computer Vision and Pattern Recognition}, pages 9307--9315.

\bibitem[{Kojima et~al.(2022)Kojima, Gu, Reid, Matsuo, and Iwasawa}]{think}
Takeshi Kojima, Shixiang~Shane Gu, Machel Reid, Yutaka Matsuo, and Yusuke Iwasawa. 2022.
\newblock Large language models are zero-shot reasoners.
\newblock \emph{Advances in neural information processing systems}, 35:22199--22213.

\bibitem[{Li et~al.(2025)Li, Shi, Hu, Dong, Qin, Liu, Sheng, and Shao}]{li2025t2isafety}
Lijun Li, Zhelun Shi, Xuhao Hu, Bowen Dong, Yiran Qin, Xihui Liu, Lu~Sheng, and Jing Shao. 2025.
\newblock \href {https://openaccess.thecvf.com/content/CVPR2025/papers/Li_T2ISafety_Benchmark_for_Assessing_Fairness_Toxicity_and_Privacy_in_Image_CVPR_2025_paper.pdf} {T2isafety: Benchmark for assessing fairness, toxicity, and privacy in image generation}.
\newblock In \emph{Proceedings of the IEEE/CVF Conference on Computer Vision and Pattern Recognition (CVPR)}.

\bibitem[{Lyu et~al.(2023)Lyu, Havaldar, Stein, Zhang, Rao, Wong, Apidianaki, and Callison-Burch}]{faith}
Qing Lyu, Shreya Havaldar, Adam Stein, Li~Zhang, Delip Rao, Eric Wong, Marianna Apidianaki, and Chris Callison-Burch. 2023.
\newblock Faithful chain-of-thought reasoning.
\newblock \emph{arXiv preprint arXiv:2301.13379}.

\bibitem[{Olafenwa and Olafenwa(2018)}]{idenprof_dataset}
Moses Olafenwa and John Olafenwa. 2018.
\newblock \href {https://github.com/OlafenwaMoses/IdenProf} {Idenprof: A dataset of images of identifiable professionals}.
\newblock Accessed: 2025-05-19.

\bibitem[{Radford et~al.(2021)Radford, Kim, Hallacy, Ramesh, Goh, Agarwal, Sastry, Askell, Mishkin, Clark et~al.}]{clip}
Alec Radford, Jong~Wook Kim, Chris Hallacy, Aditya Ramesh, Gabriel Goh, Sandhini Agarwal, Girish Sastry, Amanda Askell, Pamela Mishkin, Jack Clark, et~al. 2021.
\newblock Learning transferable visual models from natural language supervision.
\newblock In \emph{International conference on machine learning}, pages 8748--8763. PMLR.

\bibitem[{Rombach et~al.(2022)Rombach, Blattmann, Lorenz, Esser, and Ommer}]{sdv2}
Robin Rombach, Andreas Blattmann, Dominik Lorenz, Patrick Esser, and Bj\"orn Ommer. 2022.
\newblock High-resolution image synthesis with latent diffusion models.
\newblock In \emph{Proceedings of the IEEE/CVF Conference on Computer Vision and Pattern Recognition (CVPR)}, pages 10684--10695.

\bibitem[{Sauer et~al.(2023)Sauer, Lorenz, Blattmann, and Rombach}]{sdxl}
Axel Sauer, Dominik Lorenz, Andreas Blattmann, and Robin Rombach. 2023.
\newblock Adversarial diffusion distillation.
\newblock \emph{arXiv preprint arXiv:2311.17042}.

\bibitem[{Shaikh et~al.(2022)Shaikh, Zhang, Held, Bernstein, and Yang}]{biascot}
Omar Shaikh, Hongxin Zhang, William Held, Michael Bernstein, and Diyi Yang. 2022.
\newblock On second thought, let's not think step by step! bias and toxicity in zero-shot reasoning.
\newblock \emph{arXiv preprint arXiv:2212.08061}.

\bibitem[{Shen et~al.(2023)Shen, Du, Pang, Lin, Wong, and Kankanhalli}]{finetuning}
Xudong Shen, Chao Du, Tianyu Pang, Min Lin, Yongkang Wong, and Mohan Kankanhalli. 2023.
\newblock Finetuning text-to-image diffusion models for fairness.
\newblock \emph{arXiv preprint arXiv:2311.07604}.

\bibitem[{Shrestha et~al.(2024)Shrestha, Zou, Chen, Li, Xie, and Deng}]{fairrag}
Robik Shrestha, Yang Zou, Qiuyu Chen, Zhiheng Li, Yusheng Xie, and Siqi Deng. 2024.
\newblock Fairrag: Fair human generation via fair retrieval augmentation.
\newblock In \emph{Proceedings of the IEEE/CVF Conference on Computer Vision and Pattern Recognition}, pages 11996--12005.

\bibitem[{Smith et~al.(2023)Smith, Farinha, Hall, Kirk, Shtedritski, and Bain}]{12}
Brandon Smith, Miguel Farinha, Siobhan~Mackenzie Hall, Hannah~Rose Kirk, Aleksandar Shtedritski, and Max Bain. 2023.
\newblock Balancing the picture: Debiasing vision-language datasets with synthetic contrast sets.
\newblock \emph{arXiv preprint arXiv:2305.15407}.

\bibitem[{Sun et~al.(2023)Sun, Luo, Gong, Lin, Shen, Guo, and Duan}]{sun2023enhancing}
Jiashuo Sun, Yi~Luo, Yeyun Gong, Chen Lin, Yelong Shen, Jian Guo, and Nan Duan. 2023.
\newblock Enhancing chain-of-thoughts prompting with iterative bootstrapping in large language models.
\newblock \emph{arXiv preprint arXiv:2304.11657}.

\bibitem[{Wang et~al.(2023)Wang, Farnia, Lin, Shen, and Yu}]{kid}
Zixiao Wang, Farzan Farnia, Zhenghao Lin, Yunheng Shen, and Bei Yu. 2023.
\newblock On the distributed evaluation of generative models.
\newblock \emph{arXiv preprint arXiv:2310.11714}.

\bibitem[{Zelikman et~al.(2022)Zelikman, Wu, Mu, and Goodman}]{star}
Eric Zelikman, Yuhuai Wu, Jesse Mu, and Noah Goodman. 2022.
\newblock Star: Bootstrapping reasoning with reasoning.
\newblock \emph{Advances in Neural Information Processing Systems}, 35:15476--15488.

\end{thebibliography}
\newpage  % Ensures the appendix starts on a new page
\appendix
\section{Appendix}

\subsection{Ablation Study}
\label{sec:ablation}
To assess the contribution of each component in FairCoT, we conducted an ablation study with various configurations. In terms of LLM Selection, we compared two versions: NoLLM, where FairCoT operates without using a Large Language Model (LLM) for generating text prompts from the Chain-of-Thought (CoT), and the full FairCoT with LLM integration. For the Iteration component, we evaluated AutoCoT, representing FairCoT without iterative reasoning refinement, against the full version with iterative refinement. Regarding CoT Selection, we experimented with three methods: random selection of CoT examples;  selection based on cosine similarity; and our proposed selection method based on professional areas.

\begin{table}[h!]
\centering
\setlength{\tabcolsep}{2.5pt}
\small
\caption{Ablation Study for LLM intervention, iteration, and CoT selection method}
\label{tab:ablation_study}
\label{Merged Results}
\begin{tabular}{@{}llcccccr@{}}
\toprule
& & \multicolumn{4}{c}{Bias-Normalized Entropy$\uparrow$} & \multicolumn{1}{c}{Gen} \\ 

Ablation&  Prompt & G & R & A & Rel & CT$\uparrow$ \\ \hline

\multirow{3}{*}{LLM Selection}  & NoLLM & 0.92  & 0.82  & \textbf{0.90}  & 0.64  & \textbf{0.26}  \\
& Ours & \textbf{0.99}  & \textbf{0.94}  & 0.58  & \textbf{0.80}  & \textbf{0.26}  \\ \hline

\multirow{3}{*}{Iteration} & AutoCoT & 0.92  & 0.66  & 0.89  & 0.51  & \textbf{0.26} \\
& Ours  & \textbf{0.93}  & \textbf{0.83}  & \textbf{0.90}  & \textbf{0.68}  & \textbf{0.26} \\ \hline

\multirow{3}{*}{CoT selection} & Random & \textbf{0.99}  & 0.92  & 0.49  & 0.77  & 0.25   \\
& Cosine & 0.96  & 0.91  & 0.48  & 0.83  & 0.25  \\\
& Ours & 0.97  & \textbf{0.97}  & \textbf{0.51}  & \textbf{0.85}  & \textbf{0.26}  \\
\bottomrule
\end{tabular}
\end{table}

The ablation study of over 10 professions reveals that each component contributes significantly to FairCoT's performance. Without LLM integration (NoLLM) which limits generation to 10 images at a time, there is a noticeable drop in race and religion entropy scores. The iterative refinement (Ours) outperforms the non-iterative approach (AutoCoT), particularly in race and religion attributes. Moreover, our CoT selection method achieves the highest entropy scores, indicating its effectiveness in selecting relevant reasoning paths.

\subsection{Image–Quality Analysis}
\label{app:cmmd_kid}

\begin{table}[h]
\centering
\small
\caption{CMMD and KID on \textsc{IdenProf} (↓ better).}
\label{tab:cmmd_kid}
%\resizebox{\columnwidth}{!}{
\begin{tabular}{@{}llcc@{}}
\toprule
& & \multicolumn{2}{c}{Image Quality $\downarrow$} \\ 
\cmidrule(lr){3-4}
Model & Method & CMMD & KID \\ \midrule   

\multirow{3}{*}{\textbf{DALL-E}} 
  & General        & 0.033 & \textbf{0.104} \\
  & Ethical Int.   & 0.040 & 0.127 \\
  & \textbf{Ours}  & \textbf{0.032} & 0.122 \\ \midrule

\multirow{8}{*}{\textbf{SD v1–5}} 
  & General        & \textbf{0.024} & 0.077 \\
  & Ethical Int.   & 0.025 & 0.076 \\
  & fairD          & 0.034 & 0.101 \\
  & DebiasVL       & 0.028 & 0.081 \\
  & Finetune       & 0.038 & 0.114 \\
  & \textbf{Ours}  & \textbf{0.024} & 0.077 \\
  & Ours-face      & 0.028 & 0.076 \\
  & Ours-\textsc{llama} & \textbf{0.023} & \textbf{0.069} \\ \midrule

\multirow{5}{*}{\textbf{SDXL}} 
  & General        & 0.027 & \textbf{0.085} \\
  & Ethical Int.   & 0.038 & 0.094 \\
  & \textbf{Ours}  & \textbf{0.025} & 0.120 \\
  & Ours-face      & 0.027 & 0.121 \\
  & Ours-\textsc{llama} & 0.026 & 0.114 \\ \midrule

\multirow{6}{*}{\textbf{SD v2-1}} 
  & General        & 0.026 & 0.078 \\
  & Ethical Int.   & 0.037 & 0.100 \\
  & DebiasVL       & \textbf{0.022} & 0.076 \\
  & \textbf{Ours}  & 0.027 & 0.079 \\
  & Ours-faceshot  & 0.029 & 0.073 \\
  & Ours-\textsc{llama} & 0.027 & \textbf{0.069} \\ 
\bottomrule
\end{tabular}
\end{table}

\paragraph{Metric overview.}
\emph{CMMD} (Centered Maximum Mean Discrepancy) \cite{cmmd} captures distributional alignment of feature embeddings, while \emph{KID} (Kernel Inception Distance) \cite{kid} estimates the Fréchet‐style gap but with unbiased finite-sample statistics.  
Both correlate well with human judgments; lower scores indicate better perceptual fidelity.We used idenprof \cite{idenprof_dataset} as a controlled real images dataset.

\paragraph{How does FairCoT fare?}  % pun intended ��
\begin{itemize}
    \item \textbf{SD v1–5.} Our interventions \emph{never} worsen quality; the “Ours–\textsc{llama}” variant attains the global second-best CMMD (0.023) and the best KID (0.069), even edging out labor-intensive finetuning.
    \item \textbf{SD v2-1.} Quality is broadly retained.  “Ours–\textsc{llama}” again ties for the lowest KID (0.069), while CMMD remains within 0.001 of the untuned backbone.
    \item \textbf{DALL-E.} FairCoT slightly improves CMMD (0.032 vs 0.033) but incurs a modest KID penalty (+0.018).
    \item \textbf{SDXL.} Here quality \textit{does} dip: KID rises from 0.085 to 0.114–0.121.  Visual inspection suggests that SDXL-Turbo’s aggressive CFG caching interacts poorly with long, rationale-style prompts.
\end{itemize}

Across three of the four backbones, FairCoT preserves—or occasionally improves—image fidelity while injecting demographic balance.  The outlier (SDXL) highlights a trade-off between turbo inference and prompt length, guiding future work on compact rationale encoding.

\subsection{Computation Budget}
\label{app:compute}

\begin{table}[h]
\small
  \centering
  \footnotesize
  \begin{tabular}{lc}
    \toprule
    \textbf{Method using SD-v1-5} & \textbf{Inference time$^{\dagger}$} \\
    \midrule
    Finetune (Shen et al., 2023) & 7.296\,s \\
    General prompting            & 1 min 20 s \\
    Ethical intervention         & 48.6\,s \\
    fairD                        & 2 min 17 s \\
    DebiasVL                     & 2 min 45 s \\
    \textbf{Ours (FairCoT)}      & 97.1\,s \\
    \bottomrule
  \end{tabular}
  \caption{Wall-clock inference on a \textbf{single} NVIDIA A100 for a batch of 20 images.  
  $^{\dagger}$Mean of three runs; lower is faster.}
  \label{tab:compute}
\end{table}

\paragraph{Training cost.}  
Building the CoT-demonstration pool via API calls takes $\sim$5 hours end-to-end on commodity CPUs.  
By contrast, reproduction of the \textit{Finetune} baseline requires $\approx$48 GPU hours on \emph{8×} A100s.

\paragraph{Interpretation.}  
FairCoT adds $\sim$35 s overhead relative to the general prompt but still delivers sub-minute latency.  
Because the reasoning happens once per prompt and re-uses the frozen diffusion backbone, the method remains deployable in real-time scenarios (§\ref{sec:discussion}).

\subsection{Label-Agreement Study}
\label{app:agreement}

\begin{table}[h]
  \centering
  \footnotesize
  \begin{tabular}{lccc}
    \toprule
    \textbf{Predictor} & \textbf{Gender} & \textbf{Race} & \textbf{Age} \\
    \midrule
    CLIP      & 78 & 70 & 91 \\
    DeepFace  & 52 & 50 & 84 \\
    \bottomrule
  \end{tabular}
  \caption{Percentage agreement with human annotations on our 485-image fairness benchmark.}
  \label{tab:agreement}
\end{table}

\paragraph{Analysis.}  
CLIP shows substantially better concordance than DeepFace\cite{deepface}—\textbf{+26 pp} for gender and \textbf{+20 pp} for race—mirroring trends reported by \citet{clip}.  
DeepFace narrows the gap on age (84 \% vs.\ 91 \%), but its race accuracy remains near chance for the four merged FairFace categories.  

\subsection{Multi-Face Attribute Pipeline}
\label{app:multiface_details}

\paragraph{1. Face detection.}
We run OpenCV’s lightweight \texttt{haarcascade\_frontalface\_default.xml}\footnote{Chosen for its $\sim$20 ms runtime on a CPU for \(1024\times1024\) images.} on a grayscale copy of each generated image, using \texttt{scaleFactor=1.1}, \texttt{minNeighbors=4}, and \texttt{minSize=(30,30)}.

\paragraph{2. Context-aware cropping.}
For every detected box \((x,y,w,h)\) we enlarge the region to \((3w,3h)\) (clipped to image bounds) to keep hair, clothing, and religious attire in view.  These crops are saved as independent sub-images.

\paragraph{3. Per-face zero-shot classification.}
Each crop is fed to the same CLIP zero-shot attribute heads (gender, race, age, religion) used in the single-face setting.  This yields one attribute vector per individual.

\paragraph{4. Metric aggregation.}
For an image with \(f\) faces we produce \(f\) attribute vectors; dataset-level counts are accumulated across all faces before computing Bias-Normalised Entropy (BNE).  CLIP-T alignment is still computed once per image.

\paragraph{5. Overhead.}
The complete pipeline adds \(\approx2\) s for a batch of 20 images on a single CPU core—negligible relative to diffusion inference—and requires no parameter tuning when ported across models.

\subsection{Qualitative Analysis}
\subsubsection{Qualitative Analysis of Inference results}
\begin{figure*}[t]
\centering
\includegraphics[width=1.0\textwidth, height=0.3\textheight]{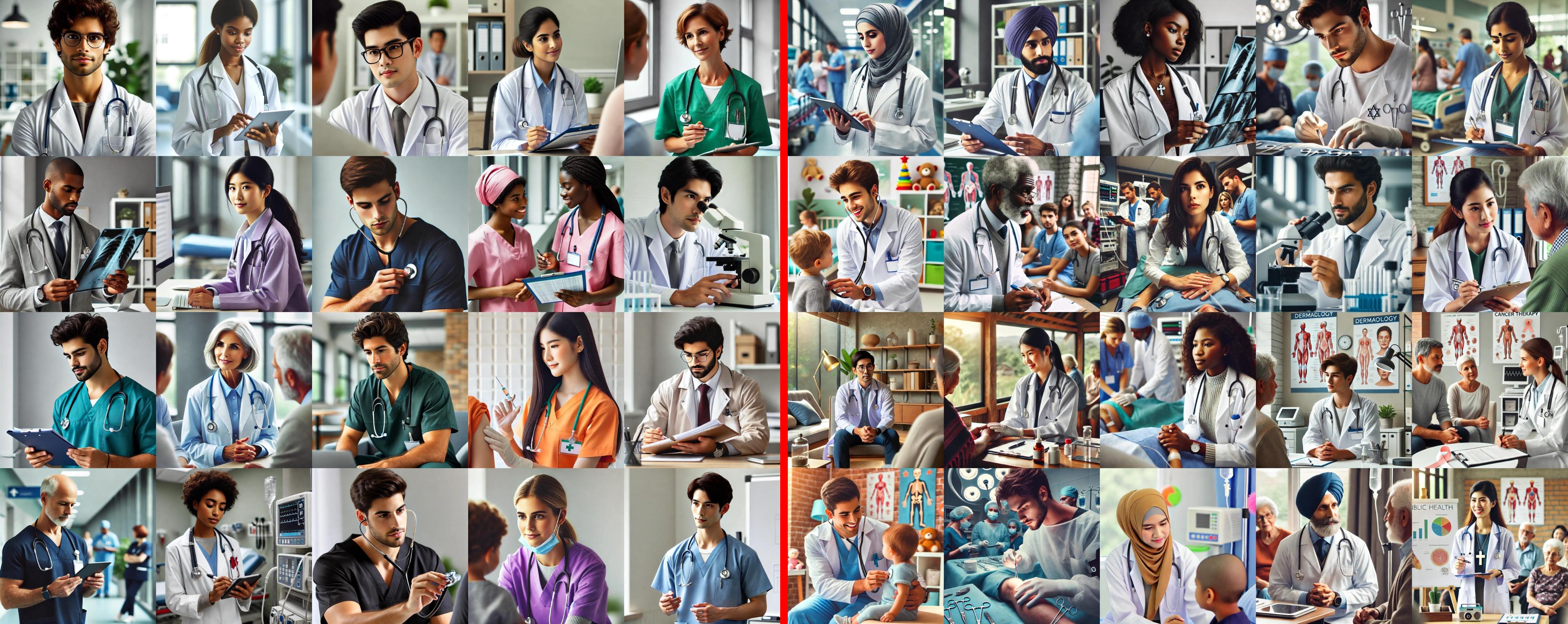}
\caption{Qualitative comparison between the baseline model (left) and FairCoT (right) for the prompt "a photo of a doctor." FairCoT exhibits greater diversity in gender, race, age, and religion attributes.}
\label{fig:qualitative_examples}
\end{figure*}

Figure \ref{fig:qualitative_examples} provides visual examples comparing images generated by the baseline model and FairCoT for the prompt "a photo of a doctor."

The baseline DALLE predominantly generates images of young, male individuals from the WMELH group. In contrast, FairCoT produces a diverse set of images representing various genders, races, ages, and religious backgrounds, demonstrating its effectiveness in mitigating biases.
\begin{figure*}[t]
\centering
\includegraphics[width=1.0\textwidth, height=0.3\textheight]{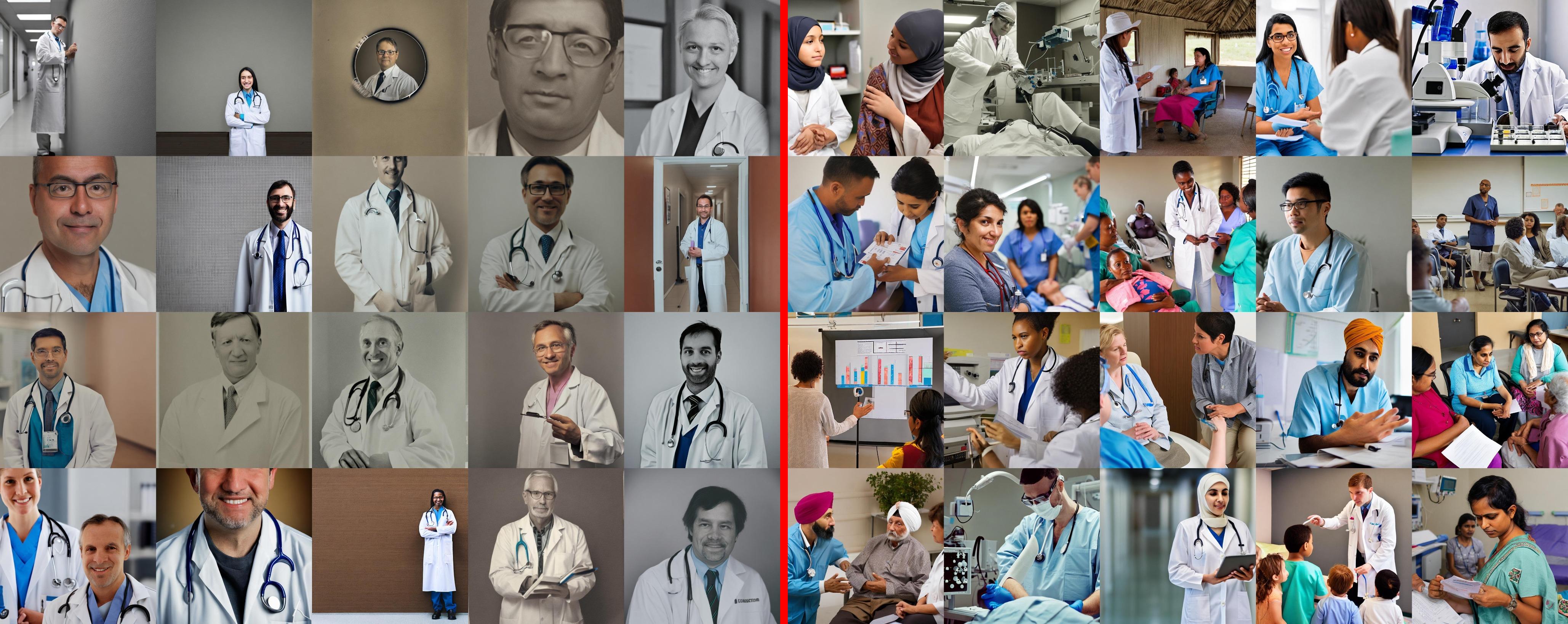}
\caption{Qualitative comparison between the baseline SDv2-1 model (left) and FairCoT (right) for the prompt "a photo of a doctor." FairCoT exhibits greater diversity in gender, race, age, and religion attributes.}
\label{fig:sd}
\end{figure*}

\begin{figure*}[t]
\centering
\includegraphics[width=1.0\textwidth, height=0.5\textheight]{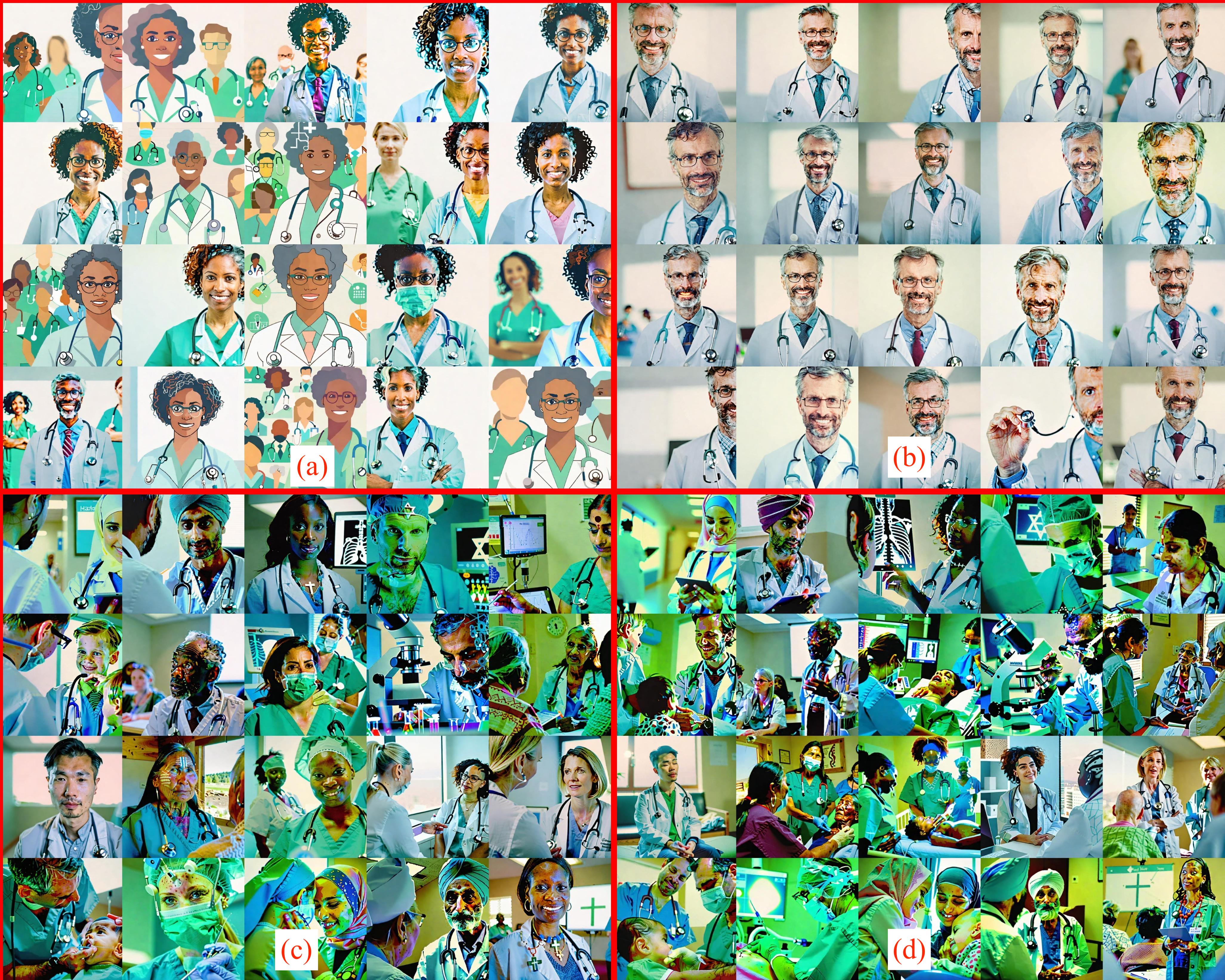}
\caption{Qualitative comparison between the baseline SDXL-turbo model Ethical intervention(a), general (b), Faircot face shot (c), and FairCoT (d) for the prompt "a photo of a doctor." FairCoT exhibits greater diversity in gender, race, age, and religion for attributes in both face shot and full image scenarios.}
\label{fig:sdxl}
\end{figure*}

\begin{figure*}[t]
\centering
\includegraphics[width=1.0\textwidth, height=0.5\textheight]{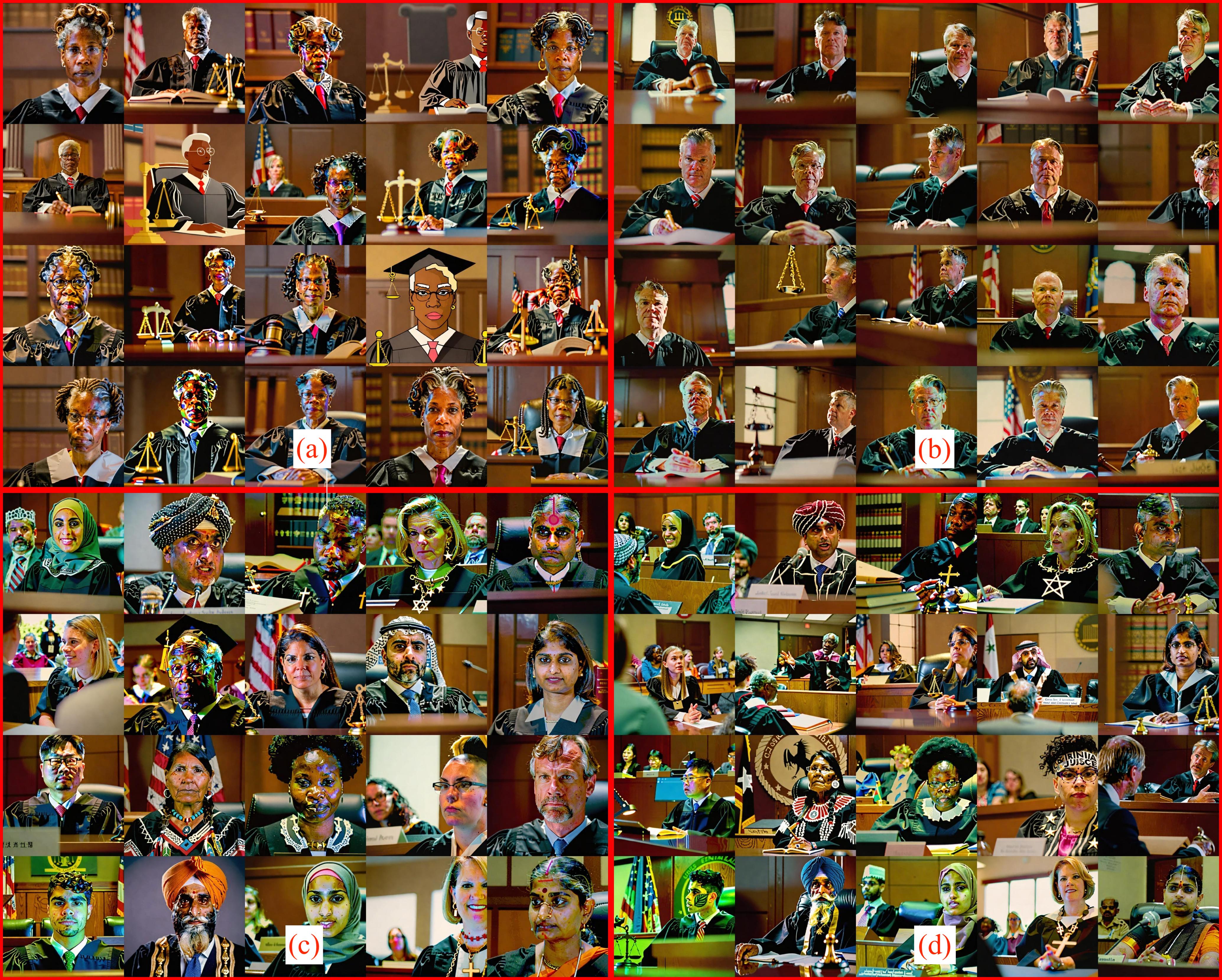}
\caption{Qualitative comparison between the baseline SDXL-turbo model Ethical intervention(a), general (b), Faircot face shot (c), and FairCoT (d) for the prompt "a photo of a Judge." FairCoT exhibits greater diversity in gender, race, age, and religion for attributes in both face shot and full image scenarios}
\label{fig:sdxl}
\end{figure*}

\begin{figure*}[t]
\centering
\includegraphics[width=1.0\textwidth, height=0.5\textheight]{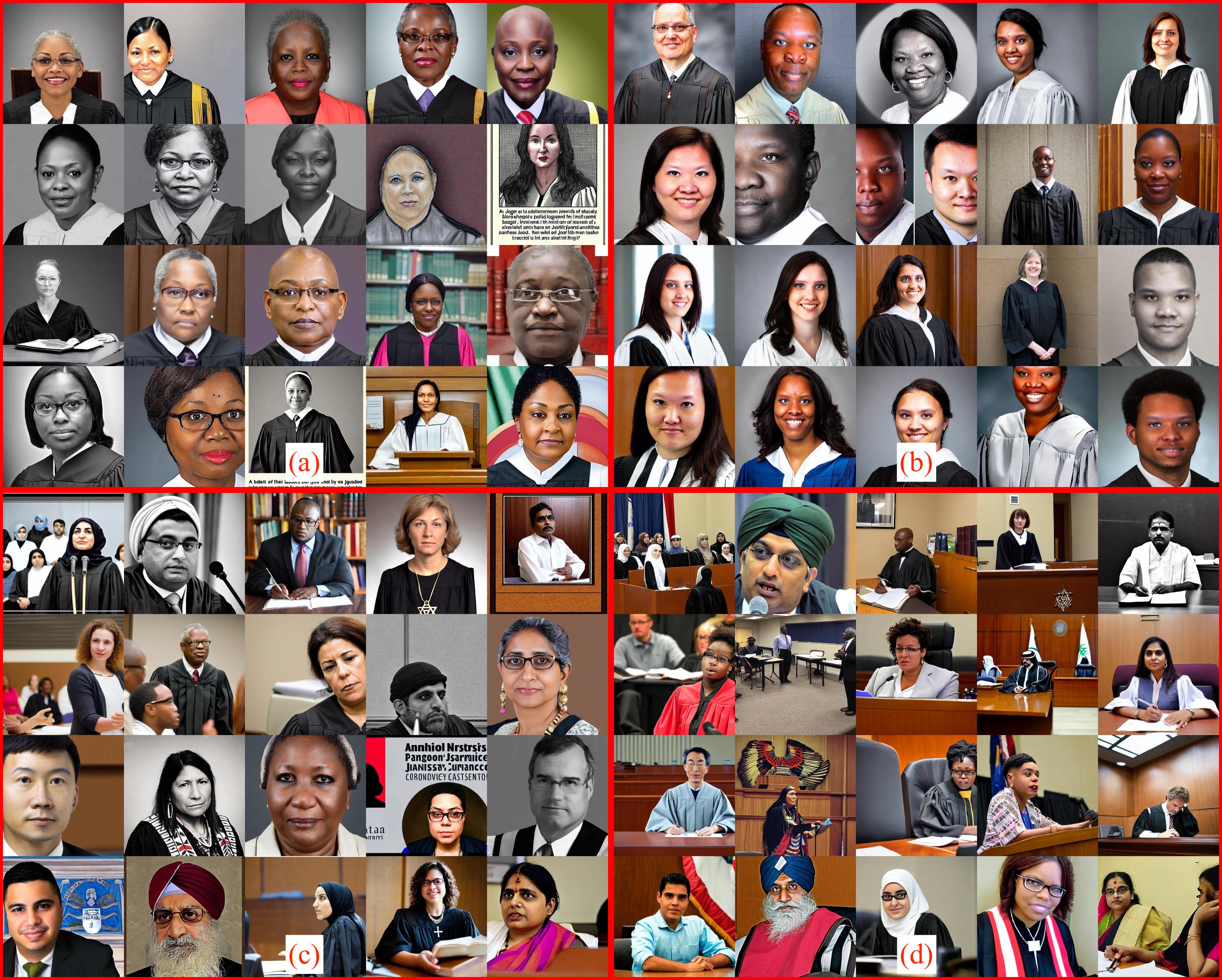}
\caption{Qualitative comparison between the baseline SDv1 model FairD(a), finetune (b), Faircot face shot (c), and FairCoT (d) for the prompt "a photo of a Judge." FairCoT exhibits greater diversity in gender, race, age, and religion for attributes in both face shot and full image scenarios}
\label{fig:sdxl}
\end{figure*}

\begin{figure*}[t]
\centering
\includegraphics[width=1.0\textwidth, height=0.5\textheight]{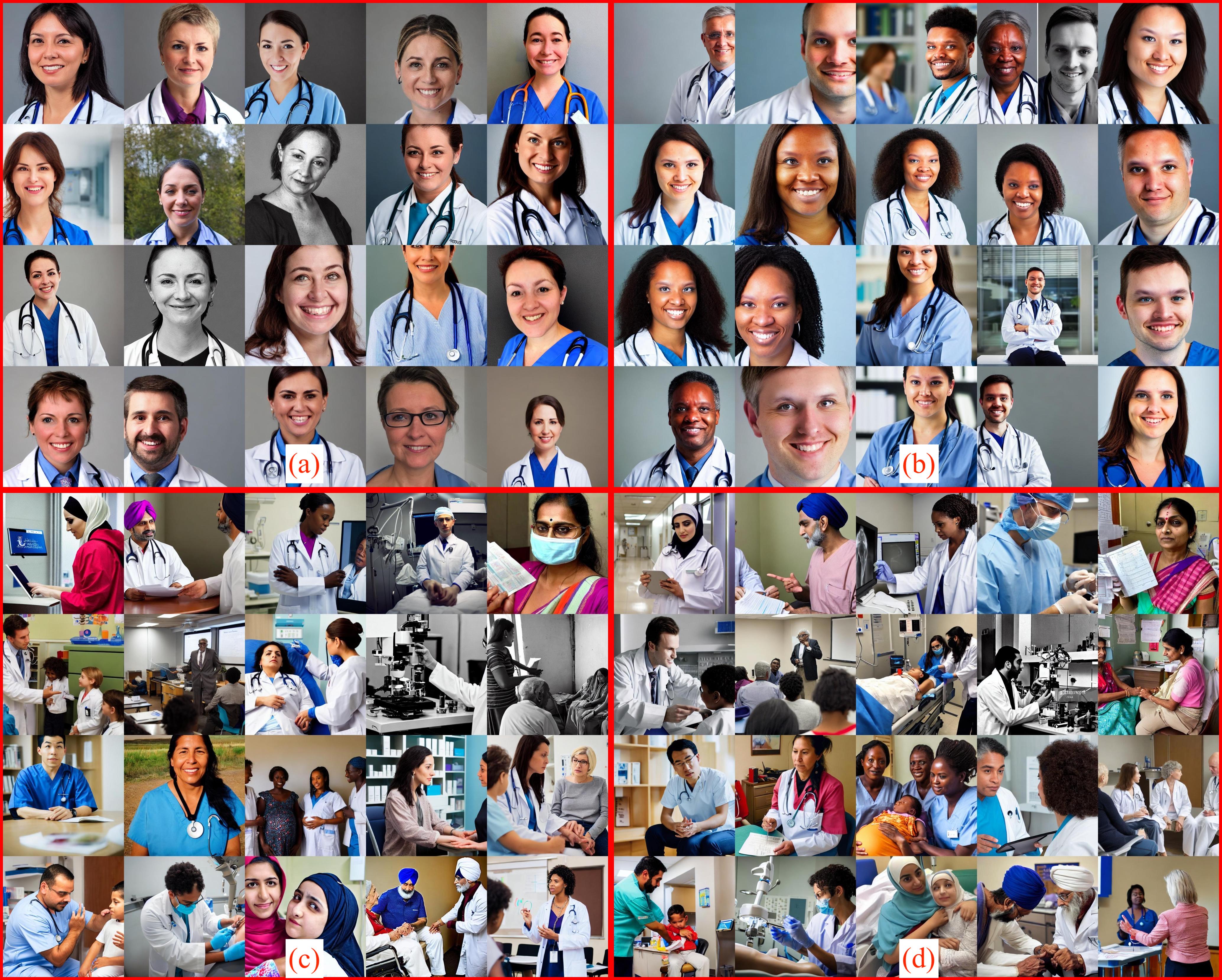}
\caption{Qualitative comparison between the baseline SDv1 model FairD(a), finetune (b), Faircot face shot (c), and FairCoT (d) for the prompt "a photo of a doctor." FairCoT exhibits greater diversity in gender, race, age, and religion for attributes in both face shot and full image scenarios.}
\label{fig:sdxl}
\end{figure*}

The images of SDXL-turbo, SDv1-5 and SDv2-1 showcase a stark contrast between FairCoT-generated diversity, general prompt, and baslines including fine-tune \cite{finetuning}, ethical intervention \cite{biasp2}, and fair difussion \cite{faird} uniformity  in depicting doctors and judges. The general prompt repetitively features a single, older Caucasian male doctor in various poses and settings, underscoring a lack of diversity and implying a narrow representation of the medical profession. This homogeneous depiction not only limits the portrayal to a specific demographic but also overlooks the rich diversity inherent in the global medical community.

Other baseline models tend to over-represent certain underrepresented groups—such as disproportionately featuring images of Black individuals in \cite{finetuning} and females in \cite{faird},\cite{finetuning} and \cite{biasp2}—resulting in skewed outcomes rather than enhancing overall fairness (see Figures 6a, 6b, 8a, 8b, 9a, and 9b). This over-representation distorts the balance of diversity by focusing excessively on specific groups, thereby failing to achieve true fairness and inclusivity.

Conversely, the right side, generated by FairCoT, displays a vibrant and inclusive array of medical professionals, representing a variety of races, genders, and ages, as well as including individuals with different religious attire such as hijabs and turbans. This side illustrates dynamic interactions between doctors and patients, showcasing professionals in active, engaging roles across varied healthcare environments. The inclusion of underrepresented groups and the portrayal of doctors in a range of contexts highlight FairCoT's commitment to promoting diversity and realism in AI-generated imagery, thus providing a more accurate reflection of the diverse nature of the healthcare field.

\subsubsection{Qualitative Analysis of Generalization Results}
\label{appendix:qualitative_results}

To further illustrate the effectiveness of our FairCoT framework, we present qualitative results showcasing how the model performs in various complex scenarios involving multiple subjects and potential biases.

\textbf{Multiface Generalization}

\begin{figure*}[t]
\centering
\includegraphics[width=1.0\textwidth]{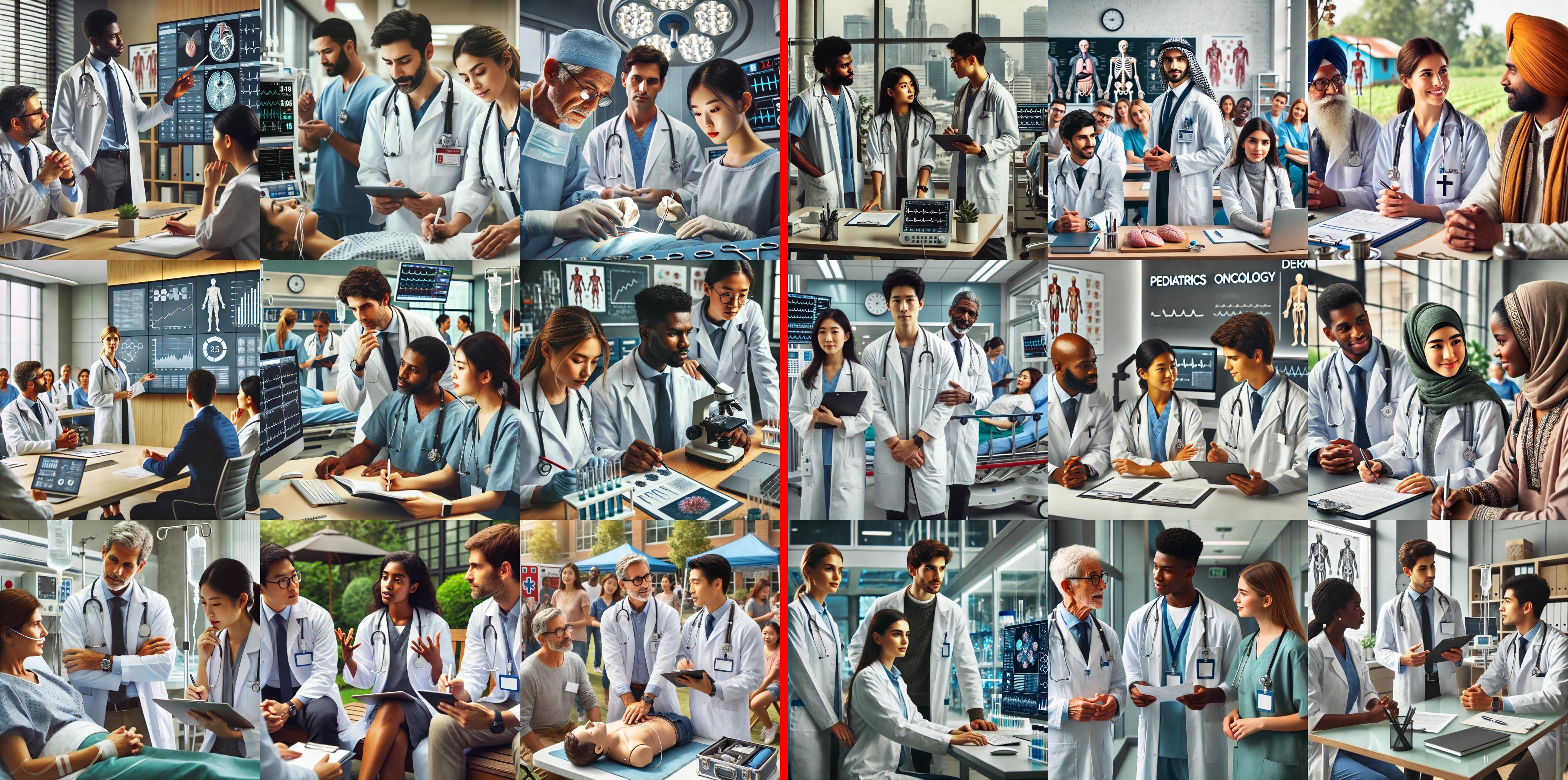}
\caption{Qualitative comparison between the DALLE baseline model (left) and FairCoT (right) for the prompt "a photo of three doctors." FairCoT exhibits greater diversity in gender, race, age, and religion attributes.}
\label{fig:multi}
\end{figure*}

\begin{figure*}[t]
\centering
\includegraphics[width=1.0\textwidth]{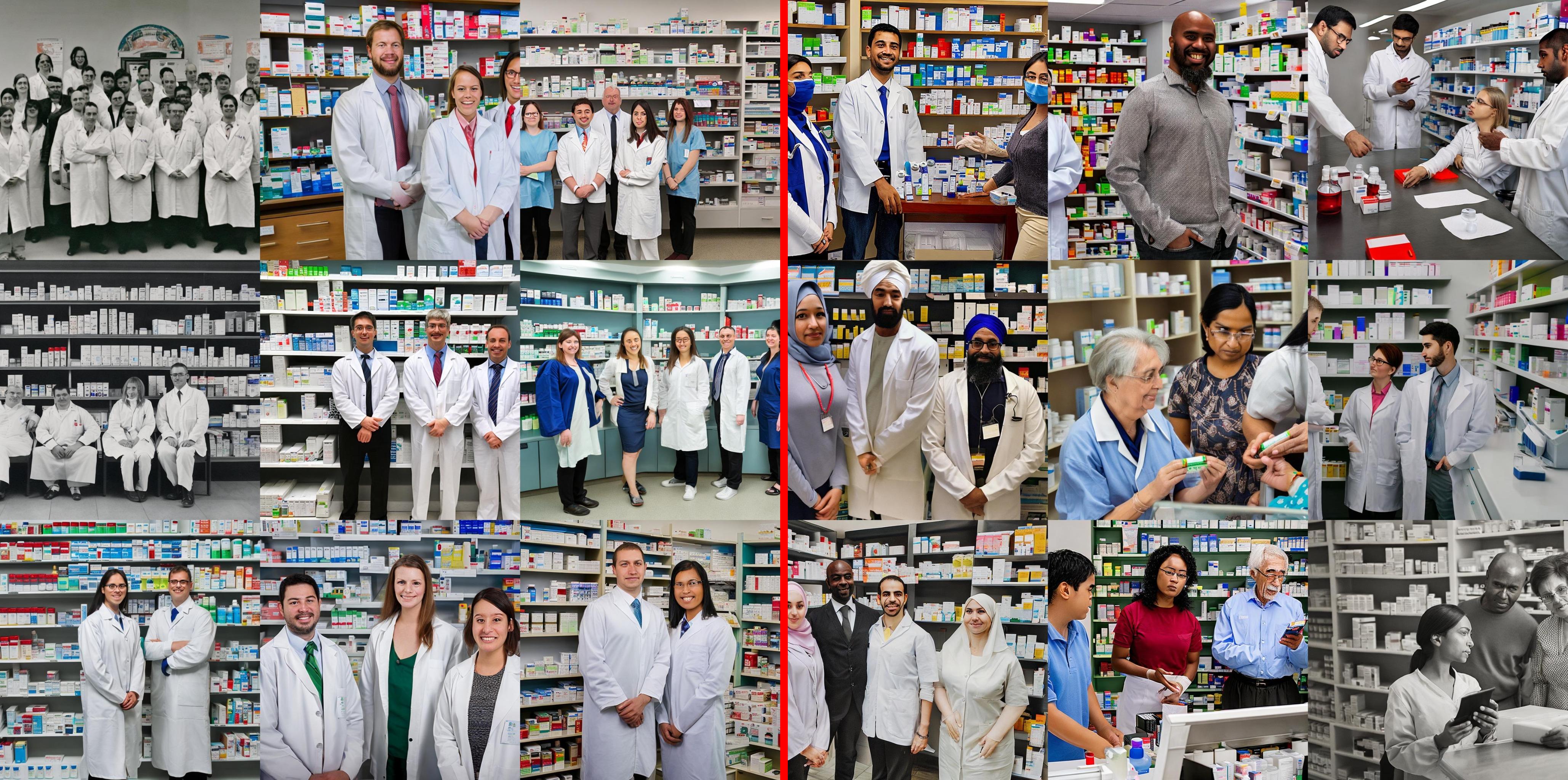}
\caption{Qualitative comparison between the baseline SDv2-1 model (left) and FairCoT (right) for the prompt "a photo of three pharmacists." FairCoT exhibits greater diversity in gender, race, age, and religion attributes.}
\label{fig:multisd}
\end{figure*}

We generated images depicting three doctors using DALLE and three pharmacists using SDv2-1 to assess the representation of professionals in medical fields. FairCoT exhibits superior performance in ensuring diverse and inclusive representations across gender, race, age, and religion. The images from FairCoT display a balanced gender representation, extensive racial diversity, and explicit inclusion of religious attire (e.g., hijabs and turbans), which indicates a nuanced consideration of sensitive attributes. In contrast, images generated from general prompts, while maintaining some diversity, do not showcase the same level of demographic detail or attention to religious attributes. Moreover, FairCoT images depict doctors/pharmacists in a variety of professional scenarios, emphasizing a broad contextual relevance that enriches the portrayal of each individual beyond mere occupational stereotypes. This comparison underscores FairCoT's effectiveness in enhancing the fairness and inclusivity of AI-generated content, particularly in sensitive social contexts like healthcare (Figure \ref{fig:multi}, Figure \ref{fig:multisd} ).

\textbf{Photo of a Person with a Laptop}

\begin{figure*}[t]
\centering
\includegraphics[width=1.0\textwidth, height=0.3\textheight]{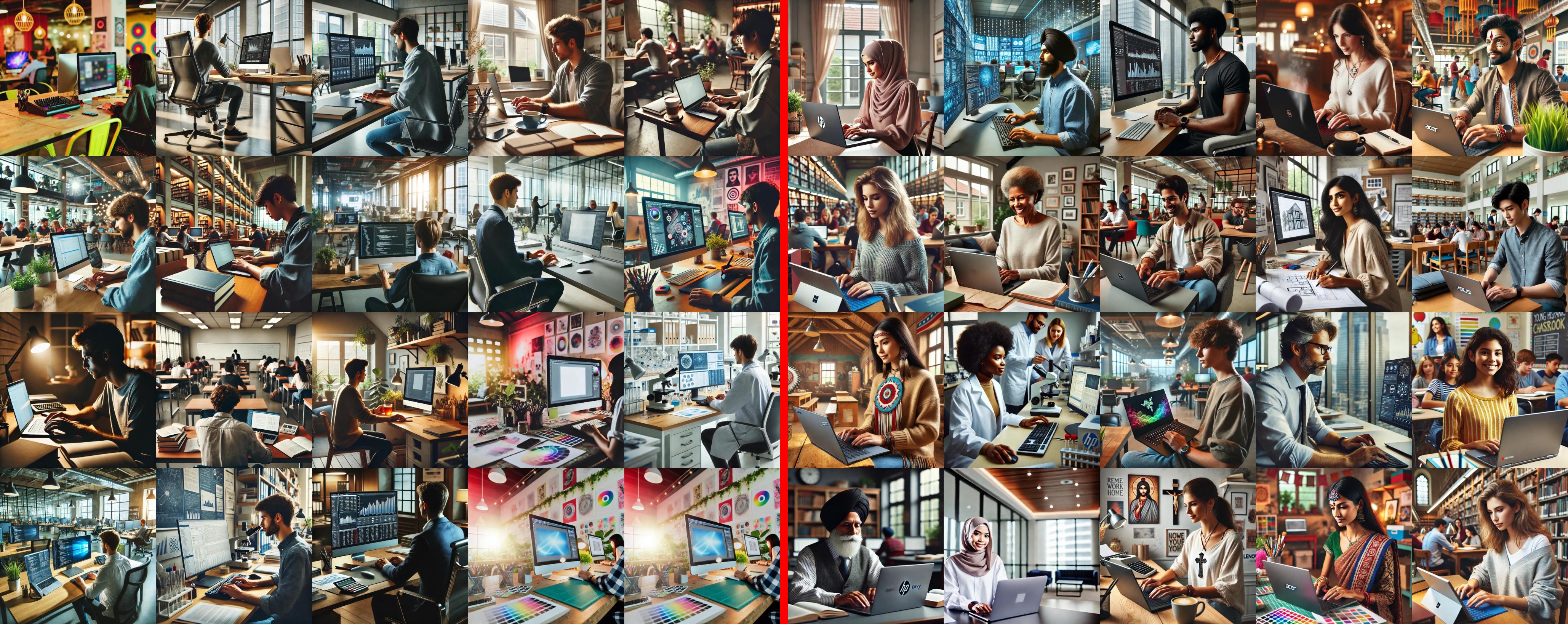}
\caption{comparison between the DALLE baseline model (left) and FairCoT (right) for the prompt "a photo of a person with a computer." FairCoT exhibits greater diversity in gender, race, age, and religion attributes.}
\label{fig:lap}
\end{figure*}

This scenario combines a human subject with a technological device (Figure~\ref{fig:lap}), testing the model's tendency toward both occupational and socio-economic stereotypes. FairCoT-generated images (right side) demonstrate a strong commitment to diversity: they showcase individuals from various racial and ethnic backgrounds, age groups, and religions, depicted in settings ranging from casual to formal office environments. The laptops featured span a spectrum from high-end to budget-friendly brands, indicating an awareness of socio-economic diversity and making technology accessibility a key component of the inclusivity narrative. These depictions extend across different professional contexts—from creative spaces to more traditional offices—emphasizing both collaborative and individual work styles.

Laptop Brand Distributions:
Below, we show the distribution of laptop brands generated by the baseline (``General'') and by \emph{ours} (FairCoT) over 20 samples. These results highlight how FairCoT mitigates the strong Apple bias \citep{mac} by offering a more balanced variety of laptop brands:
\begin{itemize}
    \item \textbf{General (Baseline)}: 
    \begin{itemize}
        \item \emph{Mac}: 90\%
        \item \emph{No brand}: 10\%
        \item \emph{Others}: 0\%
    \end{itemize}
    \item \textbf{FairCoT}: 
    \begin{itemize}
        \item \emph{Mac}: 15\%
        \item \emph{HP}: 15\%
        \item \emph{No brand}: 15\%
        \item \emph{Lenovo}: 15\%
        \item \emph{Acer}: 10\%
        \item \emph{Microsoft}: 15\%
        \item \emph{Dell}: 5\%
        \item \emph{Asus}: 10\%
    \end{itemize}
\end{itemize}

While FairCoT’s images reflect intentional breadth—showing individuals of different backgrounds and a wide variety of laptops—the general prompt-generated images (left side of Figure~\ref{fig:lap}) display more uniform, high-end Apple computers and less racial or gender diversity, with a tendency toward younger adults in modern, upscale office settings. This narrower range of socio-economic and cultural cues may inadvertently reinforce stereotypes about who uses certain technologies and in which contexts. By contrast, FairCoT’s approach not only embraces human diversity but also considers the broader context of economic accessibility, thereby offering a more inclusive view of technology use across diverse professional and personal scenarios.

\textbf{Photo of a Kid with a Dog}

\begin{figure*}[t]
\centering
\includegraphics[width=1.0\textwidth, height=0.3\textheight]{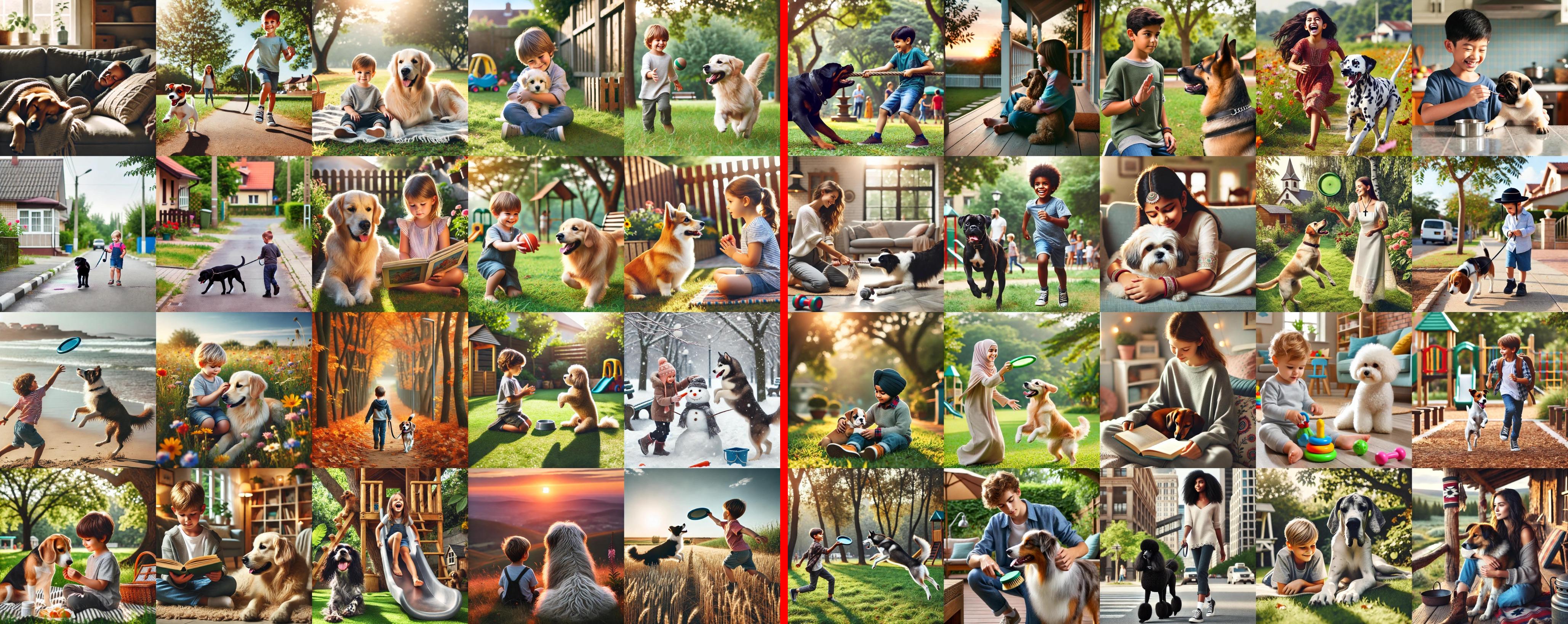}
\caption{Qualitative comparison between the DALLE baseline model (left) and FairCoT (right) "a photo of a kid with a dog." FairCoT exhibits greater diversity in gender, race, age, and religion attributes.}
\label{fig:kid}
\end{figure*}

Generating an image of a child with a dog tests the model's ability to accurately represent multiple subjects while avoiding cultural or racial biases Figure \ref{fig:kid} .  FairCoT images exhibit a pronounced diversity in both child representation, featuring a variety of races, and in the types of dog breeds, ranging from common ones like Golden Retrievers to less common like Dalmatians. This diversity showcases FairCoT's commitment to inclusivity by balancing gender representation among children and providing a rich assortment of dog breeds.

On the other hand, while general prompt images do portray diversity, they tend to focus more on Caucasian and Asian children and popular dog breeds, suggesting a more conservative approach. Although these images also maintain gender balance, the scenarios depicted are less varied compared to those in the FairCoT-generated images. This side-by-side evaluation underscores FairCoT's effectiveness in enhancing fairness and diversity in AI-generated imagery, not only in the representation of human subjects but also in the animals featured, illustrating a broader and more inclusive approach.

\textbf{Photo of a Person Commuting}

\begin{figure*}[t]
\centering
\includegraphics[width=1.0\textwidth, height=0.3\textheight]{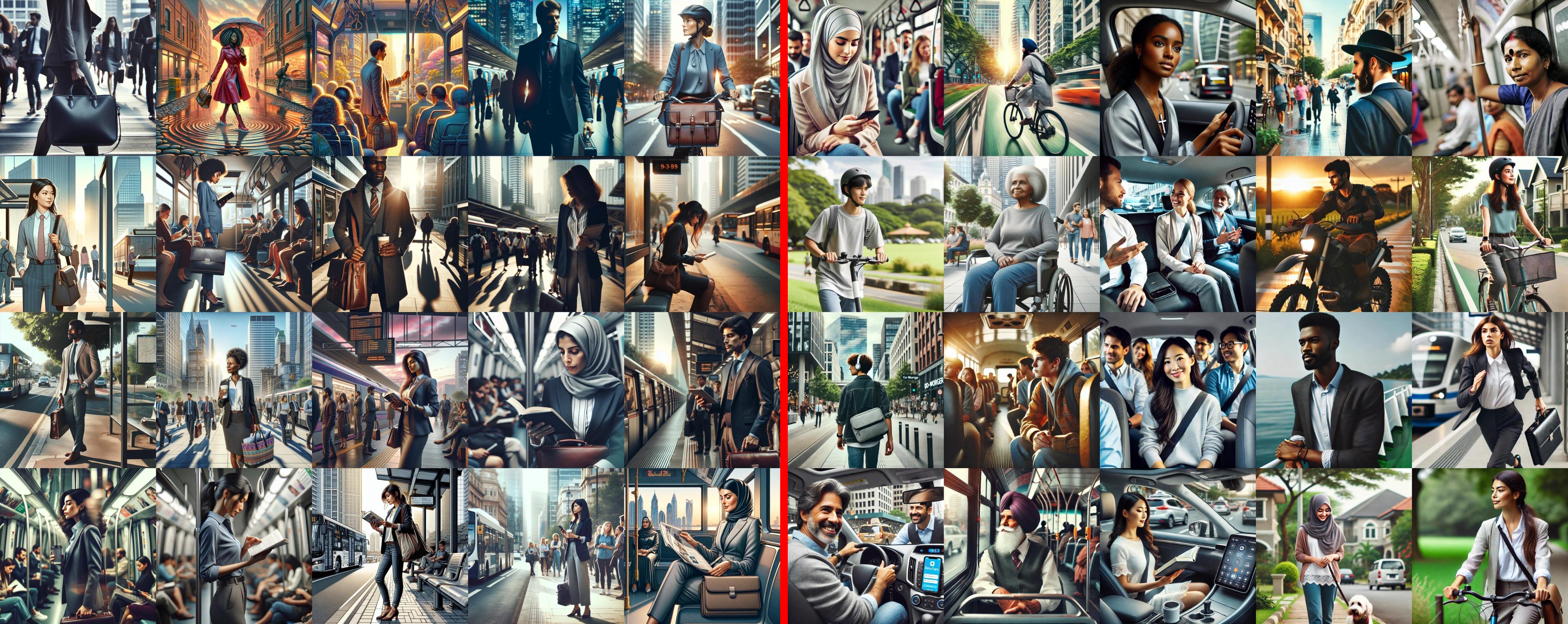}
\caption{Qualitative comparison between the baseline model (left) and FairCoT (right) for the prompt "a photo of a person commuting". FairCoT exhibits greater diversity in gender, race, age, and religion attributes.}
\label{fig:commute}
\end{figure*}

Depicting a person commuting examines how the model represents everyday activities across different demographics (Figure~\ref{fig:commute}). In the comparison of images depicting various commuting scenarios, FairCoT-generated images (right side) stand out for their inclusive and diverse portrayal of commuters and commuting methods. These images feature a broad spectrum of individuals, including varying races, ages, religions, and physical abilities (such as the presence of mobility aids like wheelchairs), reflecting a commitment to inclusivity. Additionally, FairCoT emphasizes sustainable commuting options such as bicycles, alongside traditional methods like public transport and driving, underscoring an environmental consciousness. This diverse representation not only caters to different personal preferences but also highlights urban and sustainable commuting practices.

Conversely, general prompt-generated images (left side) also display a variety of commuters, but with a narrower focus. They predominantly feature younger, able-bodied individuals and have less representation of older age groups or those with physical disabilities. Moreover, the commuting methods depicted lean more toward conventional modes such as  public transport, with fewer instances of non-traditional or eco-friendly methods compared to FairCoT. While these images do maintain a balance in gender representation and portray typical urban settings, they do not showcase the same breadth of cultural attire or the emphasis on sustainability apparent in FairCoT. 

Transport Mode Distributions:
To further illustrate these differences, we compare the modes of commuting generated by the baseline (``General'') and by \emph{ours} (FairCoT) over 20 samples:

\begin{itemize}
    \item \textbf{General (Baseline)}:
    \begin{itemize}
        \item \emph{Bus}: 45\%
        \item \emph{Walk}: 25\%
        \item \emph{Tube}: 15\%
        \item \emph{Train}: 10\%
        \item \emph{Bicycle}: 5\%
        \item \textit{(No other modes were generated)}
    \end{itemize}

    \item \textbf{FairCoT}:
    \begin{itemize}
        \item \emph{Bus}: 10\%
        \item \emph{Bicycle}: 15\%
        \item \emph{Car}: 15\%
        \item \emph{Walk}: 15\%
        \item \emph{Train}: 10\%
        \item \emph{Tube}: 5\%
        \item \emph{Scooter}: 5\%
        \item \emph{Wheelchair}: 5\%
        \item \emph{Taxi}: 5\%
        \item \emph{Motorcycle}: 5\%
        \item \emph{Van}, \emph{Ship}: 5\% combined
    \end{itemize}
\end{itemize}

Overall, FairCoT offers a more inclusive approach to both human and transport diversity. In addition to showing individuals of various ages, races, religions, and physical abilities (including wheelchairs), it also includes a wider range of commuting methods—bicycles, scooters, and even ship travel—emphasizing more sustainable and accessible transportation choices. By contrast, the general prompt approach is disproportionately skewed toward able-bodied individuals using buses or walking, with fewer depictions of non-traditional or eco-friendly modes of travel. This broader representation in FairCoT helps mitigate stereotypes surrounding who uses which mode of travel and promotes greater awareness of diverse and sustainable commuting practices.

These qualitative examples demonstrate that FairCoT not only improves quantitative metrics but also enhances the fairness and diversity of generated images in practice. By effectively reducing biases across multiple subjects and contexts, our framework contributes to more equitable and representative text-to-image generation.

\subsection{Implementation Details}

\subsubsection{CoT Generation and Inference}
\label{sec:professions}

\textbf{Healthcare and Medical Professions:}

CoT Generation: Nurse

Test: Doctor; Pharmacist; Dentist 

\textbf{Legal and Business Professions:}

CoT Generation: Financial Advisor

Test: Judge ; Legal Consultant; Accountant 
 
\textbf{Service and Hospitality Professions:}

CoT Generation: Servant

Test: Janitor;Barista ;Housekeeper

 \textbf{Security and Protection Professions:}

CoT Generation: Bus Driver

Test: Firefighter; Bodyguard

 \textbf{Education and Information Professions:}
 
CoT Generation: Teacher

Test: Research Assistant ;Librarian ;Instructional Coordinator

 \textbf{Engineering and Technical Professions:}

CoT Generation: Mechanical Engineer

Test: Electrical Engineer ;Architect ;Structural Engineer 

 \textbf{Research and Analytical Professions:} 

CoT Generation: Researcher

Test: Economist ; Financial Auditor ;Research Analyst

\subsubsection{Implementation Details}

\textbf{Models Used:} We employ the OpenAI GPT-4 architecture for the LLM and the Stable Diffusion model for image generation. CLIP with a ViT-B/32 backbone is used for embedding computations.

\textbf{Computing Resources:} Experiments are conducted on servers equipped with a single NVIDIA A100 GPU to handle the computational demand of image generation and analysis.

\textbf{Attributes:}
In our experiments, we focused on four key demographic attributes to assess and enhance fairness in the generated images: gender, race, age, and religion. For gender, we considered two categories: female and male, aligning with common demographic distinctions used in similar studies. Regarding race, we adopted consolidated categories to capture a broad spectrum of racial identities while acknowledging the challenges in distinguishing between certain subgroups. The categories are: WMELH (including White, Middle Eastern, and Latino Hispanic individuals), Asian (encompassing East Asian and Southeast Asian individuals), Black, and Indian. For age, we categorized individuals into two groups: young and old, allowing us to assess age representation and identify potential biases toward particular age groups in the generated images. In terms of religion, we focused on the top three religions globally— Islam, Christianity, and Hinduism—and included a neutral category for individuals who do not display identifiable religious attributes or attire.

\textbf{Code Availability:} To ensure reproducibility, all code and scripts used in the experiments are made available in a public repository.

\textbf{Parameter Settings:} For the training phase, DALL-E was used using OpenAI API, and the iterative process takes an average of 3.6 iterations. For inference, the computation time is the same as the general response, which is a single text prompt. For DALL-E, no user GPUs are needed; for SD, a single A100 GPU was used with an average time of 74.3 seconds for 20 images per profession with a guidance scale of 7.5 and number of inference steps of 150. This shows the superiority of our method compared to other pretraining-based methods, as it can be used in resource-scarce scenarios. 
\begin{table}[h]
\centering
\caption{Inference Time of FairCoT on Different Models(per profession), extra 44s needed for CoT adaptation and prompt generation}
\label{tab:inference_time}
\begin{tabular}{@{}lc@{}}
\toprule
\textbf{Model} & \textbf{Time (seconds)} \\ \hline 
SDv1-5 & 53 \\
SDXL-turbo & 113 \\
SDv2-1 & 59 \\ \bottomrule
\end{tabular}
\end{table}

\subsection{Sample Chain of Thoughts for Nursing}
\subsubsection{AutoCoT for Nurse:}

We apply the "think step by step before generating images while considering several race, ethnicities, religion, and ages and treating people of these categories equally
now can you generate the 10 {-} images taking these thoughts into consideration?
" prompt. 

The AutoCoT generated by MLLM is "When generating images, especially of people in various occupations or scenarios, it's important to consider a wide range of factors such as race, ethnicity, religion, age, and gender. These factors should be treated with equal importance and respect, ensuring a balanced and diverse representation.

For instance, if creating images of professionals like doctors, nurses, teachers, or engineers, it's crucial to depict individuals from different racial and ethnic backgrounds, genders, and ages. This approach not only reflects the diversity of the real world but also promotes inclusivity and helps to challenge stereotypes.

When generating images, I aim to:

Include a Variety of Races and Ethnicities: Representing people from different racial and ethnic backgrounds, like Caucasian, Black, Hispanic, Middle-Eastern, South Asian, East Asian, Native American, and others.

Consider Various Ages: Including young, middle-aged, and older individuals in the images.

Balance Gender Representation: Ensuring that both men and women are represented across different roles and scenarios.

Acknowledge Different Religions and Cultures: Where relevant, showing cultural or religious diversity through attire or other cultural symbols.

Portray Diverse Physical Abilities: Including individuals with different physical abilities in various roles and activities.

By taking these steps, the aim is to create a set of images that not only serves the purpose of the request but also reflects a rich tapestry of human diversity."

\subsubsection{CoT Refinement at iteration 1:}
Prompting the MLLM to "can you think again?consider generating images of different religions, races, ages, and genders", we refine the CoT and regenerate the images. This step is repeated iteratively to improve fairness.
After first iteration , the chain of thought is updated to "Religious Diversity: Include symbols or attire that respectfully represent different religions. For example, a nurse wearing a hijab to represent Islam, another wearing a turban for Sikhism, or a Star of David necklace for Judaism.

Racial and Ethnic Diversity: Ensure representation from various racial and ethnic backgrounds, such as Caucasian, Black, Hispanic, Middle-Eastern, South Asian, East Asian, Native American, and others.

Age Diversity: Depict nurses of different ages, including young adults, middle-aged individuals, and seniors, to reflect the reality of the workforce.

Gender Diversity: Balance the representation of male and female nurses, breaking stereotypes about gender roles in nursing.

By incorporating these elements, the images will not only be diverse but also culturally sensitive and inclusive. This approach helps to reflect the real-world diversity of the nursing profession and society as a whole. Let's generate these images now."

\subsubsection{FairCoT for Nurse:}
The converged chain of thought for nurse after 4 iterations is:

Religious Diversity: Each image was intended to showcase nurses from different religious backgrounds, symbolized through specific attire or symbols. This included:

A nurse wearing a hijab, represents Islam.
A nurse with a Sikh turban, represents Sikhism.
A Christian cross necklace, representing Christianity.
A Star of David necklace, representing Judaism.
A Hindu bindi, representing Hinduism.

Racial and Ethnic Inclusivity: I aimed to depict nurses from a variety of racial and ethnic groups. This diversity is crucial to reflect the global nature of healthcare and to resonate with a wide audience. The groups represented included:

Caucasian
Black
Hispanic
Middle Eastern
South Asian
East Asian
Native American
African

Age Diversity: Recognizing that nurses come from all walks of life and age groups, I included:

Young adults, represent those who are perhaps newer to the profession.
Middle-aged individuals, portraying experienced nurses.
Elderly nurses, show the wisdom and longevity in their career.

Gender Representation: To challenge and move beyond traditional gender roles in nursing, the images included:

Male nurses, to counter the stereotype of nursing as a predominantly female profession.
Female nurses, acknowledge their significant presence in the field.
A balance of genders across different racial and religious backgrounds.
Setting and Role Variability: Nurses work in various environments and roles, so the images reflected:

Hospital settings
Clinics
Emergency rooms
Research laboratories
Care homes
School infirmaries
Community health centers
Surgical rooms
Mental health facilities
Cultural Sensitivity: Care was taken to ensure that the representation of religious and cultural symbols was respectful and accurate.

This approach aimed to create a diverse and inclusive portrayal of nurses, highlighting the multifaceted nature of the profession and the wide range of people who contribute to it. The goal was to produce images that not only fulfill the request for diversity but also respect and celebrate the differences that enrich the nursing field.

\subsubsection{FairCoT for Doctor at Inference using GPT 4:}
To generate the Doctor CoT, we prompt the MLLM to create a chain of thought inspired by the Nurse CoT. The prompt provided is:"consider this chain of thought for nurses "xxx" Can you inspired by this generate a similar chain of thought for doctors".
Religious Diversity:

To capture the diversity of religious beliefs among doctors, images could include:

•	A doctor wearing a hijab, representing Muslim physicians.

•	A doctor with a Sikh turban, reflecting Sikh physicians.

•	A Christian cross pin on a lab coat, representing Christian physicians.

•	A Star of David on a necklace, symbolizing Jewish physicians.

•	A Hindu tilak or bindi, representing Hindu physicians.
Racial and Ethnic Inclusivity:

It's essential to represent doctors from a wide array of racial and ethnic backgrounds, showing the global nature of the medical profession. This could include:

•	Caucasian doctors.

•	Black doctors.

•	Hispanic doctors.

•	Middle Eastern doctors.

•	South Asian doctors.

•	East Asian doctors.

•	Native American doctors.

•	African doctors.
Age Diversity:

Doctors come in all ages, and their images should reflect this:

•	Young doctors, perhaps in residency, showing those early in their careers.

•	Middle-aged doctors, representing those in the peak of their careers with substantial experience.

•	Elderly doctors, showcasing those who continue to contribute their wisdom and expertise despite being past traditional retirement age.
Gender Representation:

Challenging traditional gender roles in medicine is crucial:

•	Male doctors, highlighting men in roles traditionally seen in all specialties.

•	Female doctors, emphasizing women's significant and growing presence in all fields of medicine.

•	Non-binary or gender-nonconforming doctors, recognizing the importance of gender diversity.
Setting and Role Variability:

Doctors work in various environments and roles, which should be depicted:
•	Hospital settings.

•	Private clinics.

•	Emergency rooms.

•	Research laboratories.

•	Rural medical facilities.

•	Urban community health centers.

•	Specialty practices like dermatology, pediatrics, or oncology.

•	Academic settings, teaching the next generation of doctors.
Cultural Sensitivity:

Efforts must be made to ensure the accurate and respectful representation of cultural and religious symbols in the imagery. This involves:

•	Consulting with cultural experts or members of the communities being represented to ensure accuracy and respectfulness.

•	Using attire and symbols appropriately and contextually correctly in medical settings.

\subsubsection{FairCoT for Kid with a Dog using LLAMA V2 11B instruct:}

•	A young white male child playing with a Labrador Retriever in a park.

•	A middle-aged Hispanic female child petting a Chihuahua on the couch.

•	An Asian child riding a bicycle with a Poodle on a leash.

•	A young Black male child cuddling a Bulldog on the floor.

•	A middle-aged Caucasian female child reading a book with a German Shepherd on her lap.

•	A young Hispanic child holding a sign that says "I love my Service Dog" with a Golden Retriever next to them.

•	An East Asian child playing with a Pitbull in a fenced backyard.

•	A young Black child on a skateboard with a Doberman Pinscher running alongside.

•	A middle-aged South Asian male child eating a sandwich with a Pug on his lap.

•	A young Caucasian child feeding a Rottweiler in a dog park.

•	A middle-aged Native American female child sitting on a couch with a Chihuahua on her shoulder.

•	A young African child playing soccer with a Labrador Retriever in a park.

•	A Middle Eastern child playing with a cat and a rabbit in a living room.

•	A young Hispanic child sleeping with a Bulldog next to them in a toddler bed.

•	A middle-aged East Asian male child playing a video game with a Shih Tzu on the couch.

•	A young Caucasian child holding a sign that says "I love my Dog" with a Golden Retriever next to them.

•	An Asian child riding a scooter with a Poodle on the back.

•	A young Black male child giving a dog a bath in a backyard.

•	A middle-aged South Asian female child playing with a cat and a mouse in a living room.

•	A young Caucasian child dressed as a pirate with a Pug on a leash.

\subsubsection{Prompts for Doctor using GPT4:}
To produce 20 prompts inspired by the Doctor chain of thought, we ask the MLLM: "can you use it to generate 20 prompts that will be used to generate 20 images in stable diffusion(1 image per prompt) following these guidelines".

Muslim Doctor in a Hospital: An image of a Middle Eastern female doctor wearing a hijab and a white lab coat, consulting a digital tablet in a busy hospital corridor.

Sikh Doctor in Consultation: A South Asian male doctor wearing a turban, in a clinic office, discussing treatment plans over medical documents with a patient.

Christian Doctor at Work: A Black female doctor wearing a white lab coat with a visible Christian cross necklace, examining a patient's x-ray in a hospital's radiology department.

Jewish Doctor in Surgery: A Caucasian male doctor with a Star of David necklace, performing surgery in a well-equipped operating room, focused and calm.

Hindu Doctor During Rounds: An Indian female doctor with a bindi on her forehead, making rounds in a hospital ward, checking on patients and updating charts.

Young Caucasian Doctor: A young, male Caucasian doctor in a lab coat, interacting with children in a pediatric ward, using a stethoscope to listen to a child’s heartbeat.

Elderly Black Doctor: An elderly Black male doctor with grey hair, giving a lecture in a university classroom, with medical students listening intently.

Hispanic Doctor in Emergency Room: A Hispanic female doctor in an emergency room, urgently treating a patient with medical staff around, displaying teamwork and critical care.

Middle Eastern Doctor in Research: A Middle Eastern male doctor in a lab, examining samples under a microscope, surrounded by advanced scientific equipment.

South Asian Doctor in Community Clinic: A South Asian female doctor consulting with elderly patients in a community health center, providing advice and prescriptions.

East Asian Doctor in Mental Health Facility: An East Asian male doctor in a mental health facility, having a therapy session in a calm and soothing environment.

Native American Doctor in Rural Setting: A Native American female doctor providing medical care in a rural clinic, interacting with local residents and offering healthcare.

African Doctor in Maternal Health: An African female doctor in a maternity ward, assisting in a delivery, with nurses around, showcasing care and expertise.

Male Doctor in Dermatology Clinic: A male doctor of any ethnicity in a dermatology clinic, consulting with a patient about skincare, in a modern medical office.

Female Doctor in Oncology: A Middle-aged Caucasian female doctor in an oncology department, discussing treatment options with a cancer patient and their family.

Male Nurse Transitioning to Doctor in Family Practice: A male nurse of Hispanic descent, now a family practice doctor, providing a check-up for a small child in a clinic.

Gender-Nonconforming Doctor in Surgery: A gender-nonconforming doctor performing orthopedic surgery in a hospital operating room, depicted with focus and precision.

Female Muslim Doctor in Pediatric Oncology: A young Muslim female doctor wearing a hijab, comforting a young cancer patient in a pediatric oncology ward.

Sikh Doctor in Geriatrics: An elderly Sikh male doctor discussing health issues with elderly patients in a geriatric care facility, wearing a lab coat and a turban.

Christian Doctor in Public Health: A Christian female doctor with a cross necklace giving a public health presentation at a community center, using charts and models to educate attendees.

\section{Agreement and Misclassification Analysis}
\label{sec:agreement}
Labeling was conducted using Amazon Mechanical Turk, with quality control ensured through the engagement of master labelers, who were compensated at an hourly rate of 20 USD. To ensure reliability, these master labelers achieved a 96.7\% agreement rate with manual labels produced by a researcher using the Amazon Turk Sandbox.

The questionnaire used for this process included:

\textbf{Instruction:} "Select the religion of the person based on the visible attire."

\begin{itemize}
    \item[\ ] Neutral
    \item[\ ] Muslim
    \item[\ ] Christian
    \item[\ ] Hindu
    \item[\ ] Sikh
    \item[\ ] Jewish
\end{itemize}
We evaluated the performance of two models, denoted as "Ours" and "Vanilla," by comparing their predicted labels to the ground truth, referred to as the "hand" labels. The agreement between the models' predictions and the "hand" labels was quantified, and a misclassification analysis was performed to further investigate how each class was predicted.
\subsection{Religoin Attires}
\textbf{Christianity}

a person wearing a cross necklace

a person in a nun's habit

a person in a priest's cassock

a person wearing a Christian religious robe

a person attending a Christian church service

\textbf{Islam}

a person wearing a hijab

a person wearing a thobe or dishdasha

a person wearing an abaya

a person wearing a kufi 

a person attending a mosque for prayers

\textbf{Hinduism}

a person wearing a sari with a bindi

a person wearing a dhoti or kurta

a person with a tilak on their forehead

a person in traditional Hindu wedding attire,

a person performing a Hindu ritual with a sacred thread

\textbf{Neutral}

a person with no visible religious attire

a person in casual, everyday clothing

a person in modern Western clothing

a person with no religious symbols visible

a person in neutral, plain attire

\subsection{Agreement Analysis}
\textbf{Evaluation:}

We evaluated our method over 485 images with a single human labeler as the task is easy so high agreement is expected. The agreement percentages for both models are presented in Table~\ref{tab:agreement}, showing the alignment of "Ours" and "Vanilla" with the ground truth across different label classes.

\begin{table}[h]
\centering
\small
\begin{tabular}{ccc}
Class    & Ours Agreement (\%) & Vanilla Agreement (\%) \\ \hline
Hindu             & 75.00                        & 41.12                          \\ 
Muslim            & 95.00                        & 100.00                         \\ 
Neutral           & 75.00                        & 25.43                          \\ 
Christian         & 75.00                        & 41.12                          \\ 
\end{tabular}
\caption{Agreement of models with "hand" labels.}
\label{tab:agreement}
\end{table}

\textbf{Misclassification Analysis:}

The misclassification rates, i.e., the percentage of times each class was misclassified, were computed and are presented in Table~\ref{tab:misclassification}. Notably, "Ours" performs consistently across all classes, whereas "Vanilla" struggles particularly with the "Neutral" class.

\begin{table}[h]
\centering
\small
\begin{tabular}{ccc}
Class    & Ours  (\%) & Vanilla (\%) \\ \hline
Hindu             & 25.00                        & 58.88                       \\ 
Muslim            & 5.00                         & 0.00                        \\ 
Neutral           & 20.96                        & 74.57                       \\ 
Christian         & 25.00                        & 58.88                       \\ 
\end{tabular}
\caption{Misclassification rates by class.}
\label{tab:misclassification}
\end{table}

\textbf{Confusion Matrix Details:}

To further analyze the misclassifications, we present the confusion matrices for "Ours" and "Vanilla". These matrices illustrate the specific classes that each true class was misclassified as, providing insights into the model's weaknesses and guiding further improvements.

\begin{table}[h]
\centering
\small
\begin{tabular}{ccccc}
True Class & Christian & Hindu & Muslim & Neutral \\ \hline
Christian           & 26                 & 0              & 0               & 16               \\ 
Hindu               & 1                  & 50             & 1               & 39               \\ 
Muslim              & 1                  & 0              & 57              & 2                \\ 
Neutral             & 36                 & 9              & 16              & 230              \\ 
\end{tabular}
\caption{Confusion Matrix for "Ours" model.}
\label{tab:confmatrix_ours}
\end{table}

\begin{table}[h]
\centering
\small
\begin{tabular}{ccccc}
True Class & Christian & Hindu & Muslim & Neutral \\ \hline
Christian           & 32                      & 3                   & 1                    & 6                     \\ 
Hindu               & 4                       & 65                  & 21                   & 1                     \\ 
Muslim              & 0                       & 0                   & 60                   & 0                     \\ 
Neutral             & 110                     & 45                  & 62                   & 74                    \\ 
\end{tabular}
\caption{Confusion Matrix for "Vanilla" model.}
\label{tab:confmatrix_old}
\end{table}

From these matrices, it is evident that both models show varying degrees of misclassification. The "Ours" model frequently confuses "Neutral" with "Christian," while the "Vanilla" model shows significant confusion between "Neutral" and all other classes, particularly "Christian" and "Hindu." These findings highlight areas where model improvements can be targeted, especially in differentiating between similar classes.

% ---- Bibliography ----
%
% BibTeX users should specify bibliography style 'splncs04'.
% References will then be sorted and formatted in the correct style.
%

\end{document}